\documentclass[lettersize,journal]{IEEEtran}
\usepackage{amsmath,amsfonts}
\usepackage{algorithmic}
\usepackage{algorithm}
\usepackage{array}
\usepackage{pifont}
\usepackage{multirow}
\usepackage[caption=false,font=normalsize,labelfont=sf,textfont=sf]{subfig}
\usepackage{textcomp}
\usepackage{url}
\usepackage{verbatim}
\usepackage{graphicx}
\usepackage{cite}
\usepackage{subfig}
\usepackage{gensymb}
\usepackage[dvipsnames]{xcolor}
\begin{document}

\title{FLOGA: A machine learning ready dataset, a benchmark and a novel deep learning model for burnt area mapping with Sentinel-2}

\author{Maria Sdraka, Alkinoos Dimakos, Alexandros Malounis, Zisoula Ntasiou, Konstantinos Karantzalos, Dimitrios Michail, Ioannis Papoutsis
\thanks{M. Sdraka and I. Papoutsis are with Orion Lab, Institute of Astronomy, Astrophysics, Space Applications \& Remote Sensing National Observatory of Athens, Greece.}
\thanks{M. Sdraka and D. Michail are with the Department of Informatics and Telematics, Harokopio University, Greece.}
\thanks{A. Dimakos and K. Karantzalos are with the School of Rural, Surveying and Geoinformatics Engineering, National Technical University of Athens, Greece.}
\thanks{A. Malounis and Z. Ntasiou are with the Hellenic Fire Service, Greece.}
}

\maketitle

\begin{abstract}

Over the last decade there has been an increasing frequency and intensity of wildfires across the globe, posing significant threats to human and animal lives, ecosystems, and socio-economic stability. Therefore urgent action is required to mitigate their devastating impact and safeguard Earth's natural resources. Robust Machine Learning methods combined with the abundance of high-resolution satellite imagery can provide accurate and timely mappings of the affected area in order to assess the scale of the event, identify the impacted assets and prioritize and allocate resources effectively for the proper restoration of the damaged region. In this work, we create and introduce a machine-learning ready dataset we name FLOGA (Forest wiLdfire Observations for the Greek Area). This dataset is unique as it comprises of satellite imagery acquired before and after a wildfire event, it contains information from Sentinel-2 and MODIS modalities with variable spatial and spectral resolution, and contains a large number of events where the corresponding burnt area ground truth  has been annotated by domain experts. FLOGA covers the wider region of Greece, which is characterized by a Mediterranean landscape and climatic conditions. We use FLOGA to provide a thorough comparison of multiple Machine Learning and Deep Learning algorithms for the automatic extraction of burnt areas, approached as a change detection task. We also compare the results to those obtained using standard specialized spectral indices for burnt area mapping. Finally, we propose a novel Deep Learning model, namely BAM-CD. Our benchmark results demonstrate the efficacy of the proposed technique in the automatic extraction of burnt areas, outperforming all other methods in terms of accuracy and robustness. Our dataset and code are publicly available at: https://github.com/Orion-AI-Lab/FLOGA.

\end{abstract}


\section{Introduction}
\label{sec:relatedwork}

Wildfires constitute complex global phenomena which directly affect the underlying ecosystems and can promote biodiversity and flora regeneration \cite{kelly_using_2017}. However, forest fires display a significantly increasing trend over the last decades and there is indisputable evidence that human activities are a critical factor for this \cite{portner2022climate}. Humans have been manipulating surface biomass, changing land cover schemes and oftentimes igniting fires through arson. Modern human society has also accelerated the extraction and combustion of earth’s fuels resulting in significant disturbances in the planet’s climate. Global warming changes fire regimes by cultivating the right conditions and compound effects for ignition and increasing the severity, magnitude and frequency of wildfire events \cite{senande-rivera_spatial_2022}. Such intensification of wildfires causes notable adverse consequences at multiple scales. It leads to a decline of forest stands and biodiversity, soil degradation, emission of aerosols and toxic gases, disturbance of the hydrological and carbon cycles as well as socioeconomic perils for the affected communities.

Therefore, the timely and accurate mapping of the affected area becomes an increasingly important task. Detailed and consistent information on burnt land at a fine spatial scale can greatly assist forest scientists, government agencies and local authorities to assess the damage, formulate effective response and recovery strategies as well as determine major fire drivers and implement effective fire mitigation and prevention measures where possible. In this regard, techniques such as Machine Learning (ML) and Deep Learning (DL) have the potential to leverage the abundance of remote sensing data towards automatic, robust, efficient and accurate methods for a systematic mapping of the burn scar boundary, even for inaccessible areas where field inspection is nearly impossible.

In the present study we seek to propel further research in this direction by the following contributions:

\begin{enumerate}
    \item{Benchmark dataset: A novel multi-resolution benchmark dataset is introduced which contains a great number of fire events in the wider region of Greece over the years 2017-2021. Data comprise Sentinel-2 and MODIS satellite imagery along with corresponding burnt area polygons hand-crafted by the Greek Fire Service officials, and auxiliary information such as sea masks, land cover maps, etc. We release FLOGA as open source to foster scientific research in the field.}
    \item{Thorough investigation of multiple approaches: A number of popular approaches for change detection tasks are compared and evaluated, including spectral indices, Machine Learning algorithms and Deep Learning techniques inspired by similar computer vision applications. To our knowledge, this is the first attempt in literature for such an extensive benchmarking.}
    \item{Novel Deep Learning architecture: A new Deep Learning model is proposed, BAM-CD, which achieves state-of-the-art results in the given dataset. We see potential in the developed model for application to equivalent change detection tasks in remote sensing.}
\end{enumerate}

\section{Related Work}
\label{sec:related}

\subsection{Datasets for Burnt Area Mapping}

To effectively train any Machine Learning model, it is imperative to have an extensive dataset comprising high-resolution polygons, accompanied by the corresponding satellite imagery. Ideally, such a dataset should be representative of specific ecosystems, encompass a substantial number of events, particularly given the demanding data requirements of Deep Learning models. Moreover, it should feature high-resolution satellite imagery captured before and after each event, as well as burnt area mappings with the highest achievable spatial resolution and precision.

Similar datasets are provided by several organizations as dedicated data products, most of which are relying on sensors that provide high temporal but low spatial resolution. Examples of such products are provided in Tab. \ref{tab:bam_products}. However, among those products only MTBS \cite{eidenshink_project_2007} offers mappings with high spatial resolution, i.e. in tens of meters, albeit without the corresponding satellite imagery. 

{\renewcommand{\arraystretch}{1.2}
\begin{table*}[!t]
\caption{Public and operational products for burnt area mapping.
\label{tab:bam_products}}
\centering
\resizebox{\textwidth}{!}{%
\begin{tabular}{|m{2.8cm}|m{2.4cm}|m{3cm}|m{2.5cm}|m{2.2cm}|m{2cm}|}
\hline
\textbf{Product} & \textbf{Provider} & \textbf{Satellite} & \textbf{Spatial resolution} & \textbf{Temporal compositing} & \textbf{Time span} \\
\hline
BA 300 \cite{tansey_copernicus_2017} & Copernicus & PROBA-V & 300m & 10 days & 2014-present \\
MCD64A1 \cite{giglio__louis_mcd64a1_2015} & USGS & MODIS & 500m & 1 month & 2000-present \\
FireCCI51 \cite{pettinari_m_l_esa_2020} & ESA CCI & MODIS & 250m & 1 month & 2001-2019 \\
GFed4 \cite{giglio_analysis_2013} & ORNL DAAC & ATSR, MODIS & 0.25\degree & 1 month, 1 day & 1995-present \\
GFed4s \cite{van_der_werf_global_2017} & ORNL DAAC & ATSR, MODIS & 0.25\degree & 1 month, 1 day & 1997-present \\
VNP64A1 \cite{giglio__louis_viirsnpp_2019} & USGS & VIIRS & 500m & 1 month & 2014-2019 \\
Burnt Area \cite{san-miguel-ayanz_comprehensive_2012} & EFFIS & MODIS, VIIRS & 250m, 375m & 1 day & 2007-present \\
ABoVE-FED \cite{bg-20-2785-2023} & ORNL DAAC & MODIS, Landsat & 500m & 1 year & 2001-2019 \\
MTBS \cite{eidenshink_project_2007} & USGS, USDA & Landsat & 30m & 1 year & 1984-present \\
\hline
\end{tabular}}
\end{table*}
}

To fill this gap, different researchers have assembled public datasets comprising of the burn scar polygons of several historical events. Tab. \ref{tab:bam_datasets} offers an overview of these datasets. The spatial resolution is highly improved, reaching up to 10m in most cases. Still, none of the existing datasets meet all our prerequisites for a machine learning-ready change detection formulation. This formulation necessitates an ample supply of high-resolution pre- and post-wildfire satellite imagery and their corresponding labels to serve as input data.
In particular, only the Sattelite Burnt Area Dataset, MultiEarth 2023 and HLS Burn Scar Scenes datasets include the satellite imagery corresponding to the given labels, with the latter containing only the post-event acquisitions. Among the former two, the Satellite Burnt Area Dataset contains a limited number of events in Europe (73), whereas the labels provided by MultiEarth 2023 have been obtained by FireCCI in 250m spatial resolution.

{\renewcommand{\arraystretch}{1.2}
\begin{table*}[!t]
\caption{Public datasets for burnt area mapping.\label{tab:bam_datasets}}
\centering
\resizebox{\textwidth}{!}{%
\begin{tabular}{|m{5.4cm}|m{2.2cm}|m{1.4cm}|m{2.2cm}|m{1.5cm}|m{1.4cm}|m{1.2cm}|m{1.4cm}|}
\hline
\textbf{Dataset} & \textbf{Spatial resolution} & \textbf{No. events} & \textbf{Spatial coverage} & \textbf{Time span} & \textbf{Satellite imagery included?} & \textbf{Automatic labeling?} & \textbf{Image\newline acquisitions} \\
\hline
NIFC GeoMAC Historic Perimeters \cite{noauthor_historic_nodate}, \cite{noauthor_historic_nodate2} & - & - & USA & 2000-2019 & No & -  & - \\
Satellite Burnt Area Dataset \cite{colomba_dataset_2022} & up to 10m & 73 & Europe & 2017-2019 & Yes & No & Pre, Post \\
BurnMD \cite{dotter2023burnmd} & 375m & 308 & USA & 2018-2021 & No & Hybrid & - \\
BARD \cite{franquesa_bard_2020} & up to 10m & 2661 & Global & 1988-2018 & No & Yes & - \\
MultiEarth 2023 \cite{cha2023multiearth} & up to 10m & - & Amazonia & 2001-2020 & Yes & Yes & Pre, Post \\  
HLS Burn Scar Scenes \cite{HLS_Foundation_2023} & 30m & 804 & USA & 2018-2021 & Yes & Yes & Post \\
\hline
\end{tabular}}
\end{table*}
}

\subsection{Methods for Burnt Area Mapping}
\label{ml_bam}

Traditionally, the automatic extraction of burnt areas has been primarily applied on coarse resolution satellite imagery due to the high revisit rate of such satellites, e.g. MODIS/VIIRS \cite{chuvieco_historical_2019}. Although this kind of input data provides an immediate damage assessment, it comes with a compromise on the spatial accuracy of the fire boundary and the ability to map smaller-scale fires effectively. To that end, manual refinement of the coarse mappings is often conducted by trained analysts who perform visual interpretation of higher spatial resolution satellite imagery, e.g. Sentinel-2/Landsat \cite{mitri2004semi}. However, this approach is time consuming, subject to human error and not scalable to large affected areas or multiple concurrent events. The most popular method for a fully automatic extraction of the burn scar is the computation of specific spectral indices directly on the high-resolution data. The increasing use of multispectral satellite imagery for disaster monitoring has led to the development of numerous spectral indices targeted at the identification of burnt land before, during and after a wildfire event \cite{chuvieco_historical_2019}. Such indices attempt to capture the spectral signatures of the damaged areas and are directly applied to specific sensor bands. For example, healthy, unburnt plants are characterized by high chlorophyll content thus they absorb mostly in the Red bandwidth and reflect in the NIR spectral region. Indices such as NDVI \cite{tucker1979red} and BAI \cite{Chuvieco20025103} utilize these bandwidths in order to discriminate burnt vegetation.

On the other hand, wavelengths in the visible spectrum are generally more sensitive than longer wavelength bands to atmospheric contamination such as smoke aerosols, which results in distorted measurements and ultimately erroneous assessments. For that reason, a multitude of indices (e.g. NBR \cite{key_measuring_1999} and variations, RBR \cite{keych_landscape_nodate}, MIRBI \cite{trigg_evaluation_2001}, etc) combine solely the NIR and SWIR regions, the latter of which is mostly related to leaf water content and landscape dryness \cite{huang_separability_2016}. Finally, more specialized indices have also been proposed which rely on the availability of specific satellite sensors. For instance, BAIS2 \cite{filipponi2018bais2} has been developed for the Sentinel-2 satellites and BAIM \cite{martin_performance_2005} for the MODIS constellation.

In recent years, there has been a growing interest in the use of Machine Learning (ML) techniques for the automatic extraction of burnt areas in the highest spatial and temporal resolution possible, e.g. see \cite{stavrakoudis2020automated}. These techniques have the potential to significantly improve the speed, accuracy and scalability of the entire process in a cost-effective way.
Nevertheless, while a multitude of traditional ML algorithms have been utilised for this task (see Jain et al.~ \cite{jain_review_2020} for a comprehensive review of this family of methods), DL approaches are still in their infancy. Despite the advantages of deep neural networks there are only a handful of DL techniques in the literature for tackling the task of burn scar mapping with multispectral satellite data (see Tab.~\ref{tab:bam_models} for an overview of these approaches). In particular, a number of methods take as input only post-event imagery and perform semantic segmentation where the goal is to classify each pixel in the image into one of the predefined classes of burnt/unburnt (e.g. \cite{app10124332}, \cite{knopp_deep_2020}, \cite{rs12244169}, \cite{rs14030657}, \cite{hu_uni-temporal_2021}, \cite{seydi_burnt-net_2022}). The most common models in this category employ the U-Net architecture and variations. Another approach is change detection, where the goal is to identify the changes in the landscape between two or more images taken at different times (i.e. before and after the event) (e.g. \cite{pinto2020deep}, \cite{pinto_practical_2021}, \cite{martins2022deep}, \cite{kashtan2022deep}). The most common architectures for this task is the U-Net and simple multi-layer Convolutional Neural Networks (CNNs). Finally, a single work has tackled the task through an anomaly detection approach with a CNN trained in a self-supervised way to classify a whole image patch as burnt/unburnt \cite{coca_hybrid_2022}.

\subsection{ML for Change Detection}

Change detection is a critical task in various fields, ranging from environmental monitoring to urban planning and disaster assessment. The abundance and variety of remote sensing data has provided fertile ground for the development of Machine Learning methods in change detection tasks, with Deep Learning gaining increasingly more traction over the last few years. Numerous DL architectures have emerged as powerful tools for automatically extracting hierarchical features from spatial data, and retrieving changes and transitions on the Earth's surface with high granularity \cite{cheng2023change}.

Depending on the level of detail, the methods proposed in literature can be discriminated into scene-level and region-level approaches \cite{jiang2022survey}. Scene-level approaches mainly focus on land use analysis and transitions on urban development \cite{9127795}. Region-level methods can be further divided into object-based and pixel-based change detection tasks. The former is occupied with the analysis of whole objects, which are defined as a local cluster of pixels with the same semantic label. Object-based methods are primarily applied on land use and building change monitoring, and the most common architectures include Faster-RCNN \cite{wang2018change}, Mask R-CNN \cite{rs11111343} as well as custom dual-stream models where each input image is treated separately \cite{zhang2021object}. Finally, pixel-based approaches extract features from a neighborhood of pixels and classify each individual pixel as changed or unchanged. Several studies have focused on such methods, adopting primarily fully convolutional neural networks with double-branch encoders and skip connections between the encoder and decoder levels (e.g. \cite{daudt_fully_2018}, \cite{8891676}) after the fashion of the popular U-Net model \cite{u_net}. Finally, recurrent modules have also been adopted in the literature \cite{rs12101662} to better capture temporal patterns and correlations, while some other approaches employ the Atrous Spatial Pyramid Pooling (ASPP) module \cite{chen2017rethinking} for extracting rich contextual information at multiple levels \cite{ke2021mccrnet}, \cite{ding2021dsa}.

{\renewcommand{\arraystretch}{1.2}
\begin{table*}[!t]
\caption{Related work on DL methods for burnt area mapping using multispectral data.\label{tab:bam_models}}
\centering
\resizebox{\textwidth}{!}{%
\begin{tabular}{|m{2.8cm}|m{2.4cm}|m{3cm}|m{2.5cm}|m{2.2cm}|m{2cm}|}
\hline
\textbf{Method} & \textbf{Input data} & \textbf{Input type} & \textbf{Bands} & \textbf{Output resolution} & \textbf{Open source} \\
\hline
Double-Step U-Net \cite{app10124332} & Post & S-2 (L2A) & R, NIR & 10m & \\
Knopp et al. \cite{knopp_deep_2020} & Post & S-2 (L1C) & R, G, B, NIR, SWIR & 10m & \\
Tran et al. \cite{rs12244169} & Post & UAV & R, G, B & ? & \multicolumn{1}{c|}{\ding{51}} \\
Florath et al. \cite{rs14030657} & Post & S-2 (L2A), Land cover & All & 10m & \\
Hu et al. \cite{hu_uni-temporal_2021} & Post & S-2 (L1C), L-8 & NIR, SWIR & 20m & \\
Coca et al. \cite{coca_hybrid_2022} & Post & S-2 & All & 10m & \\
Burnt-Net \cite{seydi_burnt-net_2022} & Post & S-2 (L2A) & 10m + 20m & 10m & \\
Seydi et al. \cite{SEYDI2023112961} & Post & Landsat & R, G, B, NIR, SWIR, Thermal & 30m & \multicolumn{1}{c|}{\ding{51}} \\
BA-Net \cite{pinto2020deep} & Sequence & VIIRS, FIRMS & R, NIR, MIR & 500m & \multicolumn{1}{c|}{\ding{51}}\\
Pinto et al. \cite{pinto_practical_2021} & Sequence, Pre, Post & VIIRS, FIRMS, S-2 (L1C) & R, NIR, SWIR & 10m & \multicolumn{1}{c|}{\ding{51}}\\
Martins et al. \cite{martins2022deep} & Pre, Post & L-8, PlanetScope & G, R, NIR & 3m & \\
Kashtan et al. \cite{kashtan2022deep} & Pre, Post & S-2, spectral indices & All & 10m &  \\
\hline
\end{tabular}}
\end{table*}
}

\section{Motivation}
\label{sec:motivation}

Although a number of methods customized for burnt area mapping have been proposed in the literature, most of them suffer from one or more of the following shortcomings:

\begin{itemize}
    \item{\textbf{Type of input}: Most of the aforementioned studies utilize only post-fire imagery and do not exploit any pre-fire data of the examined area which could potentially provide a valuable source of information. On the other hand, some of the proposed models require a time series of images as input which explodes the dataset size and adds an additional computational burden to the training process.}
    \item{\textbf{Evaluation process}:  The progression and behaviour of a wildfire is highly correlated with the underlying land cover type, terrain formation and local weather conditions \cite{chuvieco2019historical}. Most models are solely evaluated on a single fire event or trained and evaluated on the same event, so their generalization ability is not concretely established.}
    \item{\textbf{Manual setting of hyperparameters}: Some studies heavily rely on the manual tuning of thresholds and other hyperparameters, which hinders the usability and flexibility of the proposed method.}
    \item{\textbf{Feature engineering}: A number of approaches also define an elaborate feature extraction pipeline which somehow contradicts DL’s ability to extract useful information from raw input data.}
    \item{\textbf{Closed source}: The vast majority of existing methods are not publicly available for experimentation and testing.}
\end{itemize}

The present study provides a thorough examination of several available techniques on the problem of burnt area mapping through bitemporal satellite images, evaluated on a multitude of different fire events. We set the baselines with a number of spectral indices used traditionally in remote sensing applications and evaluate both ML and DL techniques in a vis-à-vis comparison. Additionally, we propose a novel DL architecture which achieves higher precision and outperforms current state-of-the-art approaches. Overall, the design of our research is as follows:

\begin{enumerate}
    \item{\textbf{New benchmark dataset}: A new dataset for change detection and burn scar mapping is published which includes multi-resolution satellite imagery for a large number of fire events in Greece. Ground truth labels have been manually refined at a high spatial resolution by domain experts.}
    \item{\textbf{Bitemporal input}: Both pre- and post-fire information is presented to the models in order to assist the learning process and avoid high false positive rates.} 
    \item{\textbf{Evaluation on multiple events}: The generalization ability of the examined models can be safely determined through evaluation on many different fire events that the model has not seen during training.}
    \item{\textbf{Comparison with state-of-the-art}: An application of selected state-of-the-art change detection models is presented for the task at hand and compared with our proposed approach.}
    \item{\textbf{Thorough examination of multiple approaches}: An in-depth analysis of several traditional and state-of-the-art approaches is considered in order to compose an exhaustive study which will serve as a reference guide for further improvements in the field.}
    \item{\textbf{Open source}: All code used in the present paper is based solely on open source, publicly available implementations. All code used in the development of our model is also publicly available.}
    \item{\textbf{New DL model}: We propose a novel architecture for burnt area mapping which achieves higher accuracy and better stability than off-the-shelf approaches found in literature.}
\end{enumerate}

\begin{figure*}[!t]
\captionsetup[subfloat]{captionskip=20pt}
\centering
\subfloat[]{\includegraphics[width=3.5in]{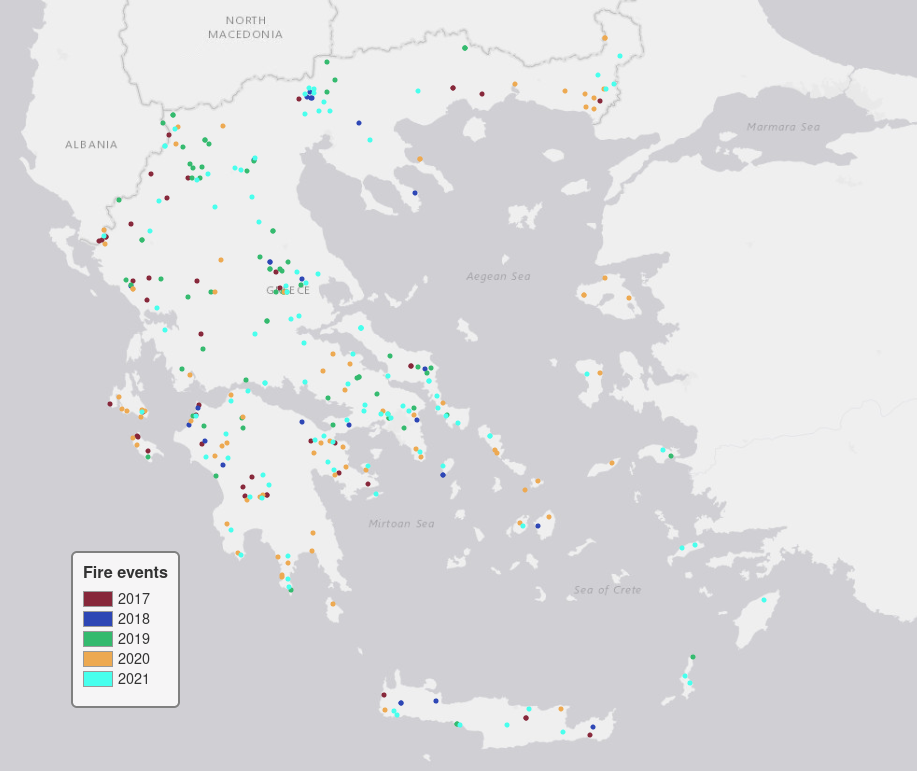}%
\label{greece_events}}
\hspace{3.5mm}%
\subfloat[]{\includegraphics[width=3.5in]{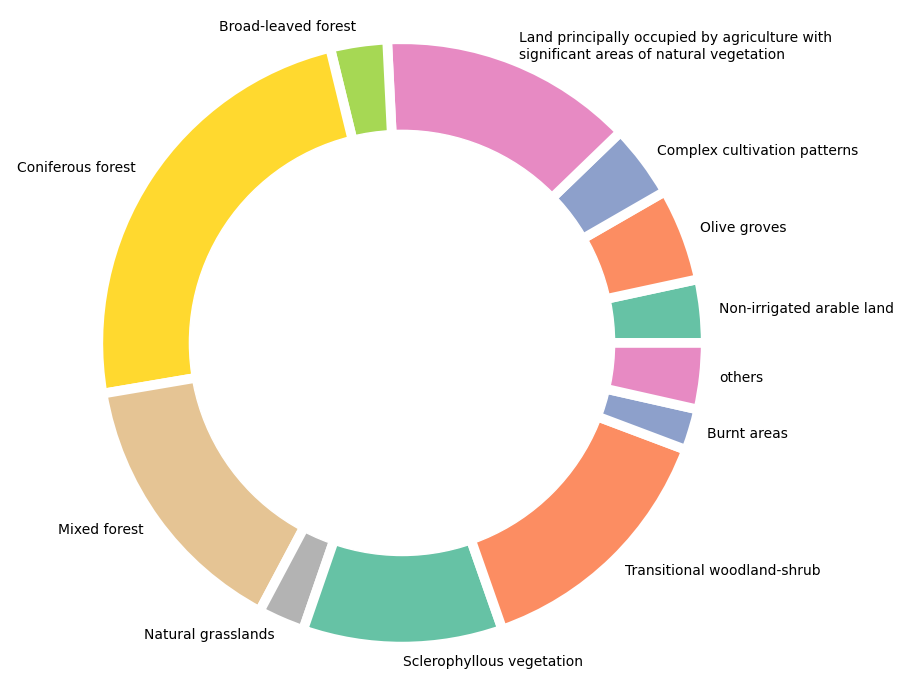}%
\label{clc_dataset}}
\caption{(a) Distribution of fire events included in our dataset, (b) Land cover types of the burnt area polygons in the dataset as defined by Corine Land Cover 2018.}
\end{figure*}

\begin{figure}[!t]
\centering
\hspace*{-1.2cm}
\includegraphics[width=3in]{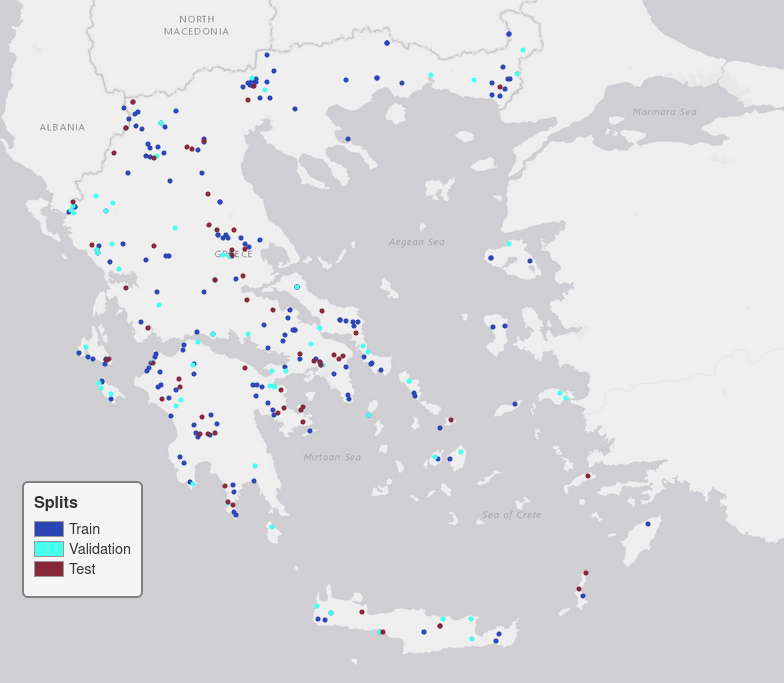}
\caption{The train/val/test split of the events.}
\label{greece_events_split}
\end{figure}

\section{Dataset}
\label{sec:dataset}

\begin{figure}[!t]
\centering
\hspace{-1.2cm}
\includegraphics[width=9cm]{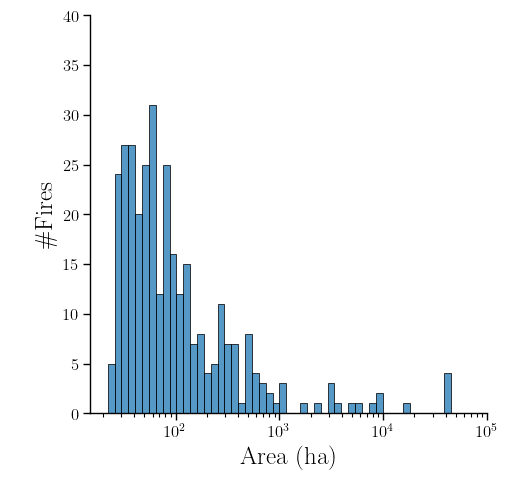}
\caption{The distribution of burnt areas per size in the FLOGA dataset. Horizontal axis is in logarithmic scale.}
\label{greece_burnt_size}
\end{figure}

We create a novel dataset of Sentinel-2 and MODIS satellite imagery acquired before and after a wildfire event, along with corresponding ground truth data which were generated semi-automatically through classification algorithms and manual photointerpretation by the Hellenic Fire Service. The dataset, named FLOGA (Forest wiLdfire Observations for the Greek Area), covers the entire region of Greece and expands over 5 years (2017-2021) including a total of 326 events.
 We have made efforts to distribute these events as evenly as possible across the geographic landscape, as illustrated in Fig. \ref{greece_events}. This approach aims to capture the diverse ecosystems across the region \cite{kokkoris2020national} and, consequently, the variations in the landscape of burnt areas \cite{2013National}. Both small and large scale events were selected, with the smallest corresponding to $\sim$22 $ha$ (0.22 $km^{2}$) and the largest to $\sim$45,284 $ha$ (452.84 $km^{2}$) (see Fig. \ref{greece_burnt_size} for a distribution of fire sizes in FLOGA). The underlying land cover categories are presented in Fig. \ref{clc_dataset} and are primarily comprising different forest types and vegetated areas. The Greek terrain is mostly mountainous and only 20.45\% of land is arable, while biomes can vary greatly according to the region and local climate. Temperate broadleaf and mixed forests can mainly be found in the northern parts of the country, Illyrian deciduous forests in the north-west, Mediterranean forests, woodlands and scrub in the region of Pindus mountain, Ionian islands and parts of Crete, and schlerophyllous and mixed forests in the central and southern part of Greece, as well as some of the Aegean islands. Based on the Köppen climate classification \cite{koppen2011}, the climate in Greece is predominantly Mediterranean with local variations and sub-climates. In general, the highest temperatures are found between mid-July and mid-August which is also usually the peak of the wildfire season that lasts from May until October. This is also reflected in the FLOGA dataset, where 55.10\% of all wildfire events occur on July and August, while 20.41\% occur on September and 11.56\% on April through June.

\begin{figure*}[!t]
\centering
\includegraphics[width=\linewidth]{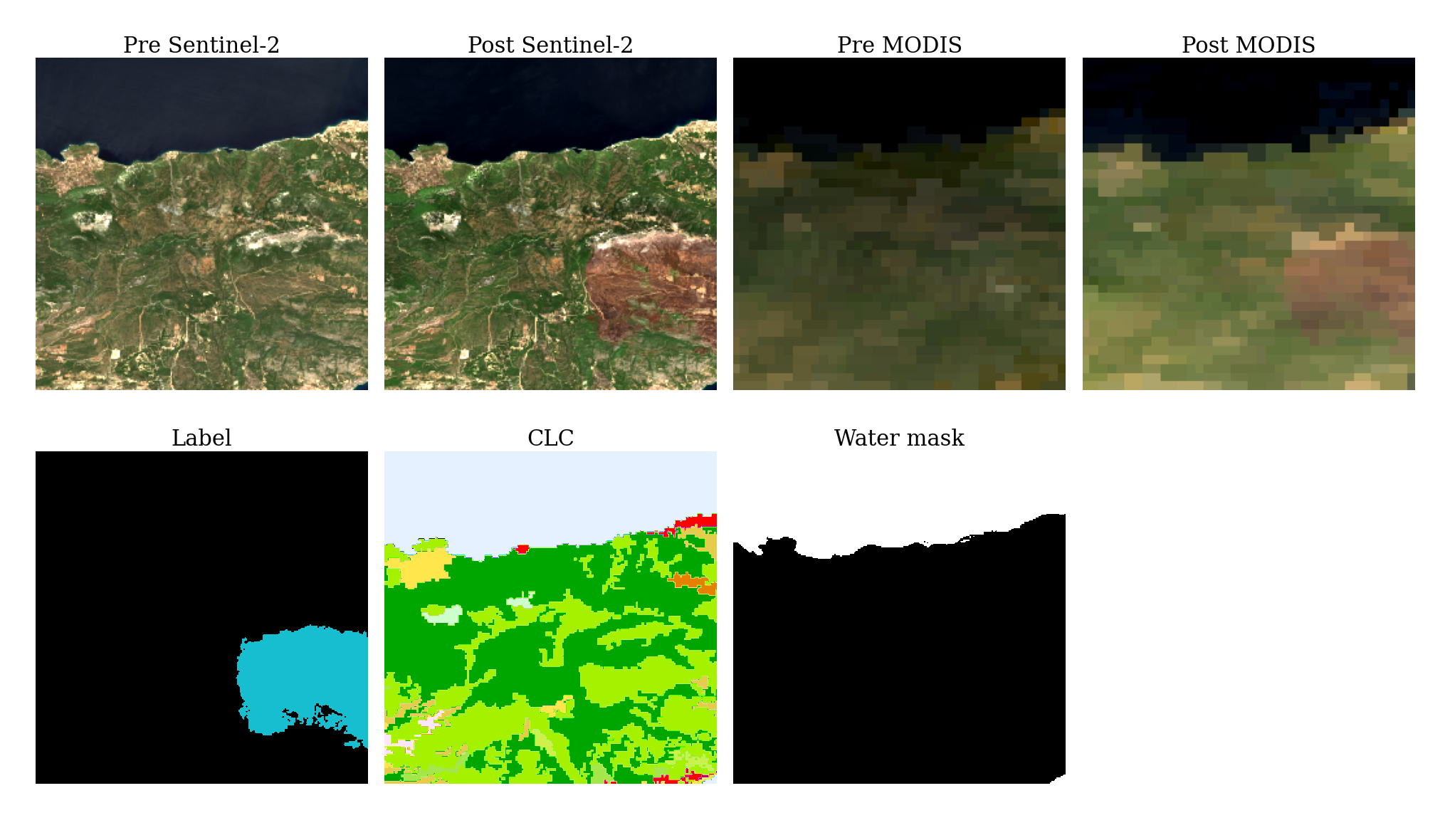}
\caption{A sample datapoint from the FLOGA dataset. We can see both Sentinel-2 and MODIS images from before and after the fire, the ground truth label, the Corine Land Cover mask (CLC) and the water mask.}
\label{data_sample}
\end{figure*}

For the satellite imagery, we only considered analysis-ready Level-2A Sentinel-2 products which are radiometrically and atmospherically corrected. The images were chosen to be dated as close to the fire event as possible and with zero cloud coverage above the area of interest. FLOGA also contains MODIS (MOD09GA) imagery for each Sentinel-2 observation, which renders it an ideal tool for multi-resolution segmentation tasks. Moreover, a high-resolution sea mask at 10m SGD and Corine Land Cover 2018 maps at 100m SGD are also provided as auxiliary information (Fig. \ref{data_sample}). To our knowledge, this is the first time that such an extensive dataset containing both imagery and labels is made available for public use.

Each Sentinel-2 image corresponding to an event is further split into 256 $\times$ 256 non-overlapping patches. Subsequently, we manually remove patches with clouds and/or cloud shadows, as well as patches containing more than 90\% of sea or water. All Sentinel-2 bands with 20m spatial resolution are considered, along with the 10m bands resampled to 20m spatial resolution. We found that clipping of pixel outliers to a predefined maximum value boosted the performance of the selected algorithms and helped ML models converge.

For the purpose of this study, we randomly split the events and keep 60\% for training, 20\% for validation and 20\% for testing in such a way that events in all three splits are uniformly distributed over Greece (Fig. \ref{greece_events_split}). We also keep data from all 5 years in all splits in order to capture the highest variability possible in vegetation status and meteorological conditions. Furthermore, to avoid the instillation of possible bias into the models, we include an equal number of patches with no burnt pixels at all, resulting in a total of 826 train patches, 189 validation patches and 152 test patches.

\section{Benchmark models}
\label{sec:experiments}

We assess the performance of three model families for mapping burnt areas within the FLOGA dataset: models reliant on spectral index thresholding, machine learning models, and deep learning models.

\subsection{Spectral indices}
\label{sec:indices}

In this study we examine a representative subset of the spectral indices proposed in the literature for burnt land mapping. The exact definitions and formulas can be found in Tab. \ref{tab:indices}. Additionally to the original single-observation formulations of the spectral indices, we define the \textit{differenced} or \textit{delta} versions of them as follows:

\begin{equation}
\label{eq:delta_index}
dSI = SI_{pre} - SI_{post}
\end{equation}

where $SI$ is any unitemporal spectral index, $SI_{pre}$ is the index value of the pre-fire image and $SI_{post}$ the index value of the post-fire image. Overall, we obtain the following differenced indices: dSAVI, dNDVI, dEVI, dNDWI, dBAI, dNBR, dNBR2, dNBR+, dMIRBI, dCSI, dBAIS2, dNBI and dABAI. 

{\renewcommand{\arraystretch}{3.0}
\begin{table*}[!t]
\caption{The spectral indices used in this study.\label{tab:indices}}
\centering
\begin{tabular}{|m{7.1cm}|m{8.5cm}|}
\hline
\textbf{Index} & \textbf{Formula}\\
\hline
Soil Adjusted Vegetation Index (SAVI) \cite{huete1988soil} & $1.5 * \cfrac{NIR - Red}{NIR + Red + 0.5}$ \\
Normalized Difference Vegetation Index (NDVI) \cite{tucker1979red} & $\cfrac{NIR - Red}{NIR + Red}$ \\
Enhanced Vegetation Index (EVI) \cite{huete2002overview} & $2.5 \cfrac{NIR - Red}{NIR + (6 * Red) - (7.5 * Blue) + 1}$\\
Normalized Difference Water Index (NDWI) \cite{mcfeeters1996use} & $\cfrac{Green - NIR}{Green + NIR}$ \\
\hline
Burned Area Index (BAI) \cite{Chuvieco20025103} & $\cfrac{1}{(0.1 - Red)^{2} + (0.06 - NIR)^{2}}$ \\
Normalized Burn Ratio (NBR) \cite{key_measuring_1999} & $\cfrac{NIR - SWIR}{NIR + SWIR}$ \\
Normalized Burn Ratio 2 (NBR2) \cite{storey2016assessing} & $\cfrac{SWIR_{1} - SWIR_{2}}{SWIR_{1} + SWIR_{2}}$ \\
Normalized Burn Ratio Plus (NBR+) \cite{alcaras2022normalized} & $\cfrac{SWIR - NIR - Green - Blue}{SWIR + NIR + Green + Blue}$ \\
Relative Differenced Normalized Burn Ratio (RdNBR) \cite{MILLER2009645} & $\cfrac{NBR_{pre} - NBR_{post}}{\sqrt{|NBR_{pre} / 1000|}}$ \\
Mid-Infrared Burn Index (MIRBI) \cite{trigg_evaluation_2001} & $(10 * SWIR_{1}) - (9.8 * SWIR_{2}) + 2$ \\
Char Soil Index (CSI) \cite{SMITH200592} & $\cfrac{NIR}{SWIR}$ \\
Burned Area Index for Sentinel2 (BAIS2) \cite{filipponi2018bais2} & $\left(1 - \sqrt{\cfrac{RedEdge * NIR_{1} * NIR_{2}}{Red}}\right) * \left(\cfrac{SWIR - NIR_{2}}{\sqrt{SWIR + NIR_{2}}} + 1\right)$ \\
Relativized Burn Ratio (RBR) \cite{rs6031827} & $\cfrac{NBR_{pre} - NBR_{post}}{NBR_{pre} + 1.001}$ \\
Normalized Burn Index (NBI) \cite{mpakairi_testing_2020} & $\cfrac{SWIR - Blue}{SWIR + Blue}$ \\
Analytical Burnt Area Index (ABAI) \cite{wu_forest_2022} & $\cfrac{(3 * SWIR_{1}) - (2 * SWIR_{2}) - (3 * Green)}{(3 * SWIR_{1}) + (2 * SWIR_{2}) + (3 * Green)}$\\
\hline
\end{tabular}
\end{table*}
}

\subsection{ML models}
\label{sec:ml_models}

Several studies have employed more traditional Machine Learning techniques to handle the task of burnt area mapping. Some of the most popular and effective approaches include the Random Forest (RF) classifier, the Support Vector Machine (SVM), the eXtreme Gradient Boosting classifier (XGBoost) and the Multilayer Perceptron (MLP). All of the aforementioned methods consider a single pixel in the input and classify it either as burnt or unburnt without combining information on the surrounding pixel neighbourhood.

For all the aforementioned ML models, we adopt a custom sampling process where a subset of the pixels contained in the original training dataset is selected in order to reduce the vast amount of candidate input data and facilitate convergence. To that end, we isolate the positive patches (i.e. patches that contain burnt land) and an equal number of negative patches (i.e. patches with no burnt land), and select $N$ number of pixels through random sampling, in such a way that $N/2$ represent burnt land and $N/2$ unburnt land. In addition, to reduce commission errors in sea/water land cover types, we also make sure to select $10\%$ of the unburnt class pixels on sea/water if present in the patch. It must be noted here that $N$ is another hyperparameter that needs to be defined at the beginning of the training process and that this sampling technique is not needed for the spectral indices or the DL models in the following section.

{\renewcommand{\arraystretch}{1.6}
\begin{table*}[ht!]
\caption{Summarization of results for the burnt area mapping algorithms. For each experiment we report the Precision (\%), Recall (\%), F1 Score (\%) and IoU (\%) for both classes as well as the mean F1 Score and the mean IoU. Numbers indicate the average value among all runs and values in parentheses indicate the standard deviation among all runs. ML results correspond to the best performing dataset among All, MI and dSI for each model. Best values are marked in \textbf{bold}, second best are \underline{underlined}.}
\label{tab:results_all}
\centering
\hspace*{-0.6cm}
\tabcolsep=0.2cm
\setlength{\arrayrulewidth}{0.2mm}
\begin{tabular}{ |m{2.3cm}||m{1.2cm} m{1.2cm} m{1.2cm} m{1.2cm}|m{1.3cm} m{1.3cm} m{1.2cm} m{1.2cm}|m{1.2cm} m{1.2cm}|| }
 \hline
  \multirow{2}{*}{\textbf{Model}}
  & \multicolumn{4}{c|}{\textbf{Unburnt}} & \multicolumn{4}{c|}{\textbf{Burnt}} & \multicolumn{2}{c||}{\textbf{Total}} \\
  \multicolumn{1}{|c||}{} & \multicolumn{1}{c}{\textbf{Precision}} & \multicolumn{1}{c}{\textbf{Recall}} & \multicolumn{1}{c}{\textbf{F1}} & \multicolumn{1}{c|}{\textbf{IoU}} & \multicolumn{1}{c}{\textbf{Precision}} & \multicolumn{1}{c}{\textbf{Recall}} & \multicolumn{1}{c}{\textbf{F1}} & \multicolumn{1}{c|}{\textbf{IoU}}& \multicolumn{1}{c}{\textbf{F1}} & \multicolumn{1}{c||}{\textbf{IoU}} \\
  \hline
  \hline
    dNBR & 98.77 & 95.31 & 97.01 & 94.19 & 34.01 & 67.12 & 45.14 & 29.15 & 71.08 & 61.67 \\
    dNBR2 & 99.13 & 95.89 & 97.49 & 95.09 & 40.12 & 76.71 & 52.78 & 35.85 & 75.13 & 65.47 \\
    dNBR+ & 98.68 & 95.97 & 97.31 & 94.75 & 36.52 & 64.27 & 46.57 & 30.35 & 71.94 & 62.55 \\
    dMIRBI & 98.95 & 98.17 & 98.56 & 97.15 & 58.3 & 71.09 & 64.06 & 47.12 & 81.31 & 72.14 \\
    RdNBR & 98.7 & 95.87 & 97.26 & 94.67 & 36.13 & 64.84 & 46.41 & 30.21 & 71.83 & 62.44 \\
    RBR & 99.02 & 94.31 & 96.61 & 93.43 & 31.91 & 74.03 & 44.6 & 28.7 & 70.6 & 61.07 \\
    dBAIS2 & 98.83 & 89.78 & 94.09 & 88.84 & 19.9 & 70.45 & 31.04 & 18.37 & 62.56 & 53.6 \\
    dNDVI & 98.84 & 95.28 & 97.02 & 94.22 & 34.45 & 68.89 & 45.93 & 29.81 & 71.48 & 62.02 \\
    dNDWI & 99.14 & 88.83 & 93.7 & 88.15 & 20.26 & 78.74 & 32.22 & 19.21 & 62.96 & 53.68 \\
    dEVI & 98.57 & 91.11 & 94.69 & 89.92 & 20.43 & 63.31 & 30.89 & 18.27 & 62.79 & 54.09 \\
    dCSI & 98.73 & 94.13 & 96.38 & 93.01 & 28.95 & 66.33 & 40.31 & 25.24 & 68.34 & 59.12 \\
    dSAVI & 98.84 & 95.28 & 97.03 & 94.23 & 34.48 & 68.89 & 45.96 & 29.83 & 71.49 & 62.03 \\
  \hline
  RF & 99.27 {\tiny (0.00)} & 97.52 {\tiny (0.01)} & 98.39 {\tiny (0.00)} & 96.83 {\tiny (0.01)} & 57.12 {\tiny (0.07)} & 82.13 {\tiny (0.02)} & 67.38 {\tiny (0.05)} & 50.08 {\tiny (0.05)} & 82.88 {\tiny (0.02)} & 73.82 {\tiny (0.03)} \\
   XGBoost & 99.42 {\tiny (0.0)} & 96.24 {\tiny (0.0)} & 97.80 {\tiny (0.0)} & 95.70 {\tiny (0.0)} & 47.88 {\tiny (0.0)} & 85.93 {\tiny (0.0)} & 61.49 {\tiny (0.0)} & 44.40 {\tiny (0.0)} & 79.65 {\tiny (0.0)} & 70.05 {\tiny (0.0)} \\
  SVM & 98.73 {\tiny (0.00)} & 98.79 {\tiny (0.00)} & 98.76 {\tiny (0.00)} & 97.55 {\tiny (0.00)} & 69.45 {\tiny (0.00} & 68.42 {\tiny (0.00)} & 68.93 {\tiny (0.00)} & 52.59 {\tiny (0.00)} & 83.85 {\tiny (0.00)} & 75.07 {\tiny (0.00)} \\
  MLP & 99.39 {\tiny (0.00)} & 95.31 {\tiny (0.00)} & 97.31 {\tiny (0.00)} & 94.75 {\tiny (0.00)} & 42.26 {\tiny (0.01)} & 85.36 {\tiny (0.01)} & 56.52 {\tiny (0.01)} & 39.4 {\tiny (0.01)} & 76.91 {\tiny (0.00)} & 67.08 {\tiny (0.00)} \\
  \hline
  FC-EF-Diff & 98.76 {\tiny (0.36)} & 99.36 {\tiny (0.6)} & 99.06 {\tiny (0.13)} & 98.13 {\tiny (0.26)} & 82.71 {\tiny (12.87)} & 65.27 {\tiny (10.32)} & 71.2 {\tiny (2.04)} & 55.32 {\tiny (2.42)} & 85.13 {\tiny (1.09)} & 76.72 {\tiny (1.34)} \\
  FC-EF-Conc & 99.01 {\tiny (0.08)} & 98.94 {\tiny (0.46)} & 98.98 {\tiny (0.2)} & 97.98 {\tiny (0.4)} & 72.47 {\tiny (8.84)} & 72.67 {\tiny (2.37)} & 72.18 {\tiny (3.6)} & 56.59 {\tiny (4.44)} & 85.58 {\tiny (1.9)} & 77.28 {\tiny (2.42)} \\
  SNUNet-CD & \underline{99.59 {\tiny (0.09)}} & 99.41 {\tiny (0.12)} & \underline{99.5 {\tiny (0.05)}} & \underline{99.01 {\tiny (0.1)}} & 84.6 {\tiny (2.46)} & \underline{88.67 {\tiny (2.48)}} & \underline{86.53 {\tiny (1.21)}} & \underline{76.28 {\tiny (1.88)}} & \underline{93.02 {\tiny (0.63)}} & \underline{87.64 {\tiny (0.99)}} \\
  ADHR-CDNet & 99.52 {\tiny (0.06)} & 99.25 {\tiny (0.17)} & 99.38 {\tiny (0.07)} & 98.77 {\tiny (0.13)} & 80.74 {\tiny (3.16)} & 86.77 {\tiny (1.81)} & 83.58 {\tiny (1.33)} & 71.81 {\tiny (1.96)} & 91.48 {\tiny (0.7)} & 85.29 {\tiny (1.05)} \\
  ChangeFormer & 99.49 {\tiny (0.07)} & \textbf{99.47 {\tiny (0.08)}} & 99.48 {\tiny (0.03)} & 98.97 {\tiny (0.07)} & \underline{85.45 {\tiny (1.74)}} & 85.9 {\tiny (2.07)} & 85.65 {\tiny (0.97)} & 74.91 {\tiny (1.48)} & 92.56 {\tiny (0.5)} & 86.94 {\tiny (0.77)} \\
  \hline
  BAM-CD & \textbf{99.7 {\tiny (0.04)}} & \underline{99.44 {\tiny (0.05)}} & \textbf{99.57 {\tiny (0.01)}} & \textbf{99.14 {\tiny (0.01)}} & \textbf{85.63 {\tiny (0.97)}} & \textbf{91.56 {\tiny (1.21)}} & \textbf{88.48 {\tiny (0.16)}} & \textbf{79.34 {\tiny (0.26)}} & \textbf{94.03 {\tiny (0.08)}} & \textbf{89.24 {\tiny (0.14)}} \\
 \hline
\end{tabular}
\vspace*{5mm}
\end{table*}
}

\subsection{Off-the-shelf DL models}
\label{sec:dl_models}

As we discussed in section \ref{ml_bam}, most of the approaches proposed in literature for burnt area mapping are not well fitted to the task at hand since they make use of unitemporal data or a sequence of multiple input images. The only methods that require bitemporal input are either not offered as publicly open implementations or utilize a small subset of the image bands, thus limiting the source of available information to the model. Therefore, we resort to more generic DL models proposed for other change detection tasks and select a wide range of methods with varying capacities and capabilities. 

FC-EF-Diff and FC-EF-Conc \cite{daudt_fully_2018} are among the first DL models proposed for change detection in Remote Sensing data. They comprise simple U-Net structures where the contracting part is a Siamese two-branch encoder with shared weights. The feature maps extracted by the two encoders are either differenced (FC-EF-Diff) or concatenated (FC-EF-Conc) at each level and then passed to the decoder through skip connections.

SNUNet-CD \cite{9355573} is a densely connected U-Net++ \cite{zhou2019unet++} architecture with a Siamese encoder whose intermediate features are concatenated before passing to the decoders through the dense skip connections. A custom attention module is also designed which takes as input all decoder outputs to produce the final mapping. This connectivity scheme aims to improve the representation of the extracted features and alleviate the loss of high-frequency information taking place in sparsely connected standard U-Net architectures.

ADHR-CDNet \cite{zhang_adhr-cdnet_2022} is a modern, state-of-the-art model proposed recently for the detection of changes in high-resolution Remote Sensing imagery. It consists of a backbone which is a modified version of HRNet \cite{wang2020deep} customized to extract multi-scale differential information from the two input images. The extracted feature maps are then fed to a multi-scale spatial feature attention module which is able to detect changes of varied scales and locations.

ChangeFormer \cite{bandara_transformer-based_2022} exploits a cascade of transformer modules in a Siamese encoder in order to achieve more accurate feature extraction and stronger context modelling. The intermediate features are differenced and passed through a lightweight MLP decoder which produces the final change map.

All of the above models take as input the 20m bands of Sentinel-2, i.e. B02, B03, B04, B05, B06, B07, B11, B12, B8A. The selection of optimizer and learning rate has been directly based on the respective publications, and each model is trained for a total of 250 epochs.

\begin{figure*}[!t]
\centering
\includegraphics[width=5.7in]{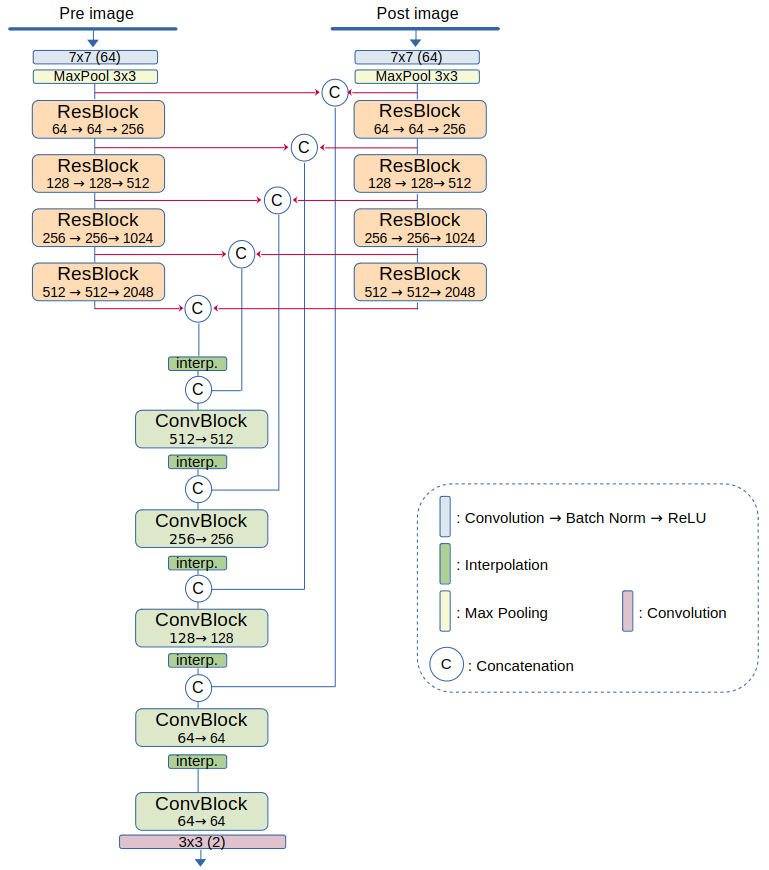}%
\caption{Architecture of the BAM-CD model. Numbers inside the ResBlock and ConvBlock modules indicate the number of output channels for each internal convolutional layer. A detailed overview of ResBlock and ConvBlock is given in Fig. \ref{modules}.}
\label{arch}
\end{figure*}

\section{BAM-CD: Our proposed model}
\label{sec:methods}

\begin{figure*}[!t]
\centering
\subfloat[]{\includegraphics[width=3.2in]{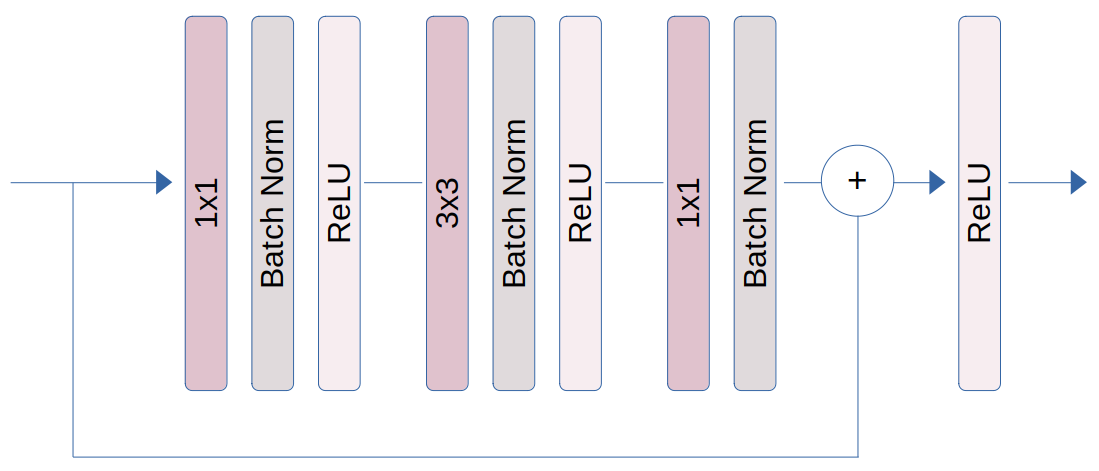}%
\label{archs_resblock}}
\hfil
\subfloat[]{\includegraphics[width=3.2in]{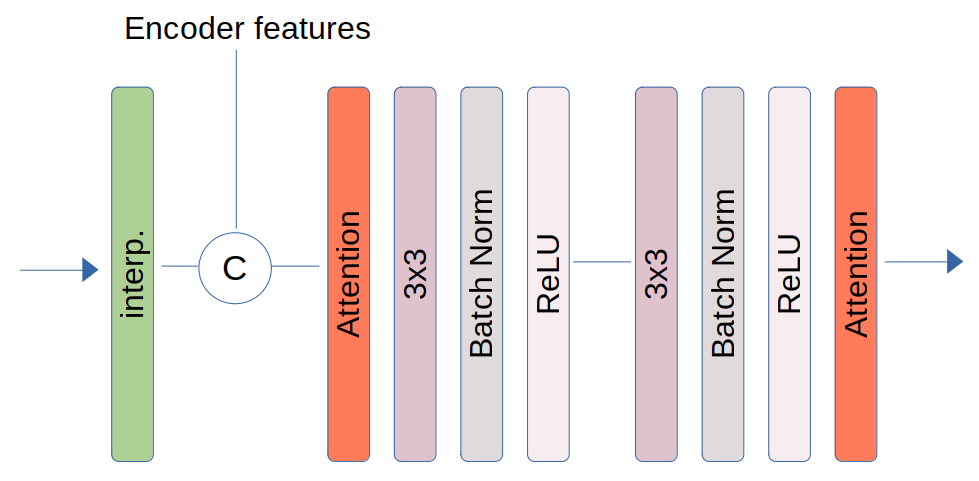}%
\label{archs_decoderblock}}
\caption{(a) The ResBlock in the BAM-CD architecture, (b) The ConvBlock in the BAM-CD architecture. The attention mechanism is shown in Fig. \ref{modules_attn} in more detail.}
\label{modules}
\end{figure*}

\begin{figure*}[!t]
\centering
\includegraphics[width=4.5in]{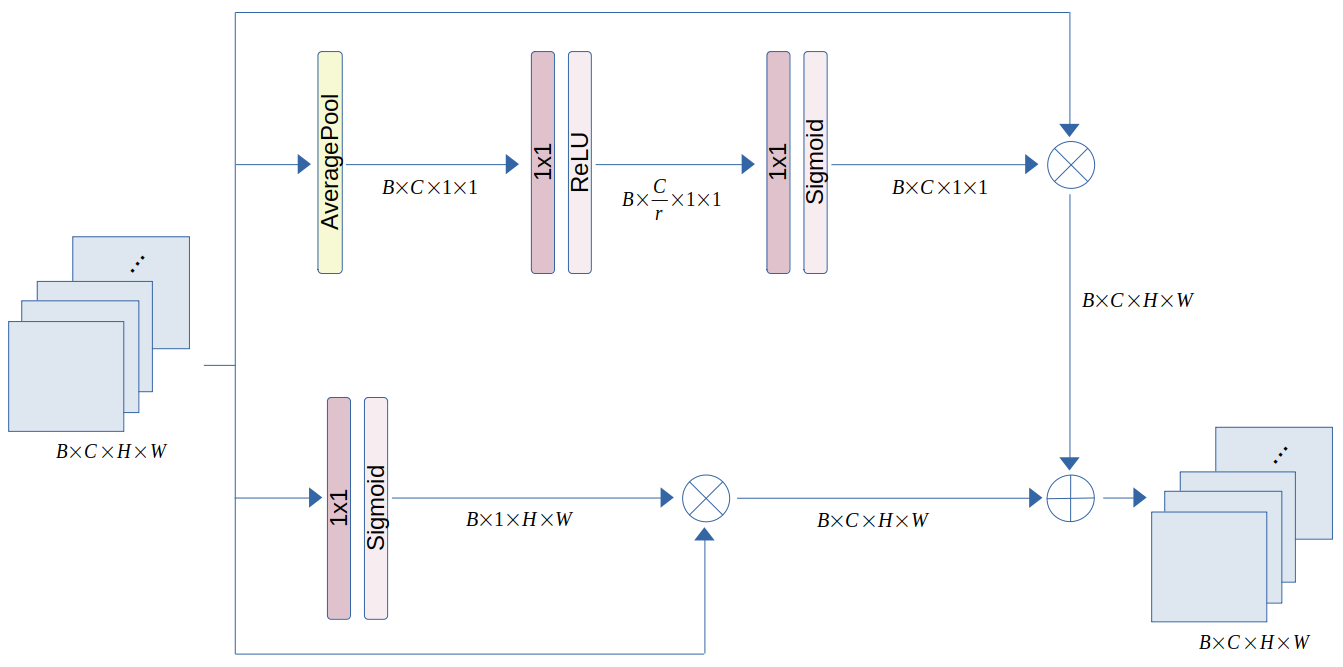}%
\caption{Schematic of the attention mechanism in the ConvBlock.}
\label{modules_attn}
\end{figure*}

We examine the problem through the prism of change detection rather than semantic segmentation in order to (i) ensure that no past burnt areas are mistakenly identified and (ii) to reduce the risk of confusion with spectrally similar surfaces (e.g. water bodies, dark soil, agricultural harvesting, etc.) (\cite{huang_separability_2016}, \cite{giglio_collection_2018}). In our approach we draw inspiration from one of the most popular architectural designs in change detection models and adopt a double-stream framework \cite{cheng2023change} where the bitemporal images are treated in parallel by two deep feature encoders. These encoders follow the DeepResUnet \cite{yi_semantic_2019} structure and are trained with a Siamese scheme meaning that their weights are shared and are trained simultaneously. Such double-stream approaches have consistently displayed better feature extraction capabilities than single-stream architectures where the images are concatenated before being fed to the model. Subsequently, the feature maps extracted by each residual block (ResBlock, Fig. \ref{archs_resblock}) are concatenated and fed to a decoder which outputs the final binary mask. The decoder comprises a cascade of convolutional blocks (ConvBlock, Fig. \ref{archs_decoderblock}) with attention modules \cite{roy2018recalibrating} (Fig. \ref{modules_attn}) which help the model focus on the most salient features and isolate the important changes in the input pair of images. The attention modules are an improved version of the popular ``Squeeze-and-Excitation'' technique \cite{hu2018squeeze} where channel and spatial dimensions are treated separately and have shown impressive results in image segmentation tasks. Finally, to take full advantage of all the intermediate features extracted in the encoders, we adopt skip connections between each encoder level and the corresponding decoder level, similar to a U-Net architecture. We call our proposed model \textit{BAM-CD} and a schematic overview of the network can be found in Fig. \ref{arch}.

After a series of experiments and testing of several variations, we select the ResNet-101 \cite{he2016deep} backbone for the image encoders and keep all 10m and 20m Sentinel-2 bands in the input, resampled to 20m resolution. BAM-CD has a total of $\sim$83.7 M trainable parameters and is trained with the Adam optimizer \cite{kingma2014adam} and a learning rate of 0.001 without any scheduling or weight decay.

\section{Experiments and results}
\label{sec:results}

To assess the efficacy and robustness of the proposed model, we perform a thorough comparison with different popular approaches for burnt area mapping. First, we explore a number of spectral indices tailored for burnt scar discovery and then we experiment with four traditional ML methods including Random Forest (RF), eXtreme Gradient Boosting (XGBoost), Support Vector Machine (SVM) and Multilayer Perceptron (MLP). Finally, we select a few representative DL models found in the literature to get state-of-the-art results and prove that BAM-CD achieves superior performance.

All of the aforementioned methods are evaluated with Precision, Recall, Intersection over Union or Jaccard Index (IoU) and F1-Score metrics. All experiments are repeated 5 times and the mean and standard deviation of the metrics are reported.

\begin{equation}
Precision = \frac{TP}{TP + FP}
\end{equation}

\begin{equation}
Recall = \frac{TP}{TP + FN}
\end{equation}

\begin{equation}
F1~Score = 2 \times \frac{Precision \times Recall}{Precision + Recall}
\end{equation}

\begin{equation}
IoU = \frac{TP}{TP + FP + FN}
\end{equation}

where $TP$ are the true positives, $FP$ the false positives and $FN$ the false negatives. In each experiment, for the final evaluation we select the best model based on the F1 score performance for the burnt class in the validation set.

\subsection{Examination of spectral indices}
\label{sec:indices_results}

In the first part of the benchmark we focus primarily on the differenced indices since we are taking a change detection approach and utilize bitemporal input imagery. Spectral indices take continuous values which describe the extend of the examined property (i.e. vegetation health, soil dryness, charred flora, etc.) for a specific pixel. Therefore, one should convert these continuous-valued indicators into a binary mapping in order to extract the burnt area. Various studies have proposed elaborate or dynamic thresholding techniques, but we presume that the derivation of a single global threshold per index is an adequate baseline for our needs and can be easily generalized. To that end, we perform a thresholding-based segmentation where the continuous values are discretized in the following manner:

\begin{equation}
\label{eq: index_thresholding}
dSI_{thresh} = \left\{
\begin{array}{ll}
      0, & dSI < threshold \\
      1, & dSI \geq threshold\\
\end{array} 
\right. 
\end{equation}

In order to find the appropriate threshold for each index, we conduct a grid search over multiple candidate threshold values and assess their performance in the train set. For the final evaluation, the best thresholds found in the previous step were applied to the test set and the final metrics were computed. Nevertheless, we found that such global thresholds are usually not easy to determine, because some spectral indices are highly sensitive to atmospheric conditions, seasonal phenology, solar angle, etc. In addition, indices such as NBI were designed to be used as input to classification algorithms (e.g. Random Forest) and not as standalone indicators. In the upper part of Tab. \ref{tab:results_all} we report the best performing indices where an adequate global threshold was found.

dMIRBI achieves superior results in most metrics, which is consistent to findings of several previous studies (e.g. \cite{lu_evaluation_2016}, \cite{fornacca_evaluating_2018}) and showcases the robustness of this index in variable land cover types. Additionally, indices non specific to burnt area discovery, such as dNDVI and dSAVI, seem to achieve quite good results for this task. This is probably due to the Red and NIR bands involved in their calculation and which have been traditionally regarded as among the most salient bandwidths for burnt land imprint.

\begin{figure}[!ht]
\centering
\hspace*{-0.8cm}
\includegraphics[width=9.5cm]{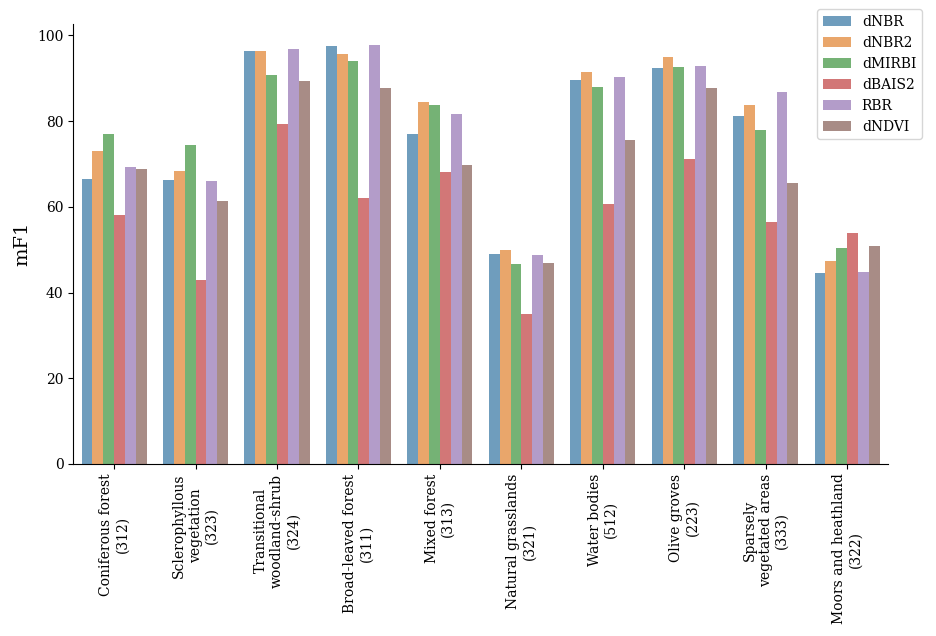}
\caption{The mean F1-score of the most robust spectral indices for selected land cover types. Numbers in parentheses indicate the corresponding Corine Land Cover code.}
\label{lc_stats}
\end{figure}

From the low scores achieved on the burnt class it is evident that all indices suffer from a high rate of commission errors. This phenomenon can be attributed to various causes which have been extensively discussed in literature. Firstly, burnt areas of minor severity often provide too weak a spectral signature for the indices to capture, especially when there is a relatively big time interval between the end of the wildfire and the acquisition of the post-event imagery \cite{pacheco_analysis_2023}. Secondly, our dataset covers the whole region of Greece, thus land cover types, biomes and vegetation structures can vary greatly among the events. Several studies have shown that spectral indices respond differently to different land cover types and thresholds are usually not transferable from one to another \cite{2013National}, \cite{rs10111680}, \cite{huang_separability_2016}, \cite{perez_evaluating_2022}. As shown in Fig. \ref{lc_stats}, there is no single index that shows superior performance for all of the selected land cover types which renders the selection of a universal index for the task considerably hard. It is interesting to note that all indices consistently display lower accuracies in moors, heathland and natural grasslands, which may be attributed to the fact that most wildfires occur during the dry summer season, thus the reflectance spectra of such dry, non-photosynthetic vegetation is quite similar to the spectra of burnt post-fire vegetation or exposed soil.
Thirdly, the time interval between the event and the captured imagery plays a crucial role. Different vegetation types display different regeneration patterns and indices have been found quite sensitive to the time lag of the image acquisition \cite{fornacca_evaluating_2018}. Lastly, most spectral indices are heavily affected by atmospheric and meteorological conditions, shadows from clouds or topography variations, as well as the presence of water bodies, all of which may display similar spectral signatures with burnt land \cite{wu_forest_2022}, \cite{alcaras2022normalized}. Such an effect is shown in the selected FLOGA samples of Fig. \ref{test_vis_3}, where a large number of false positives are predicted on the sea, and Fig. \ref{test_vis_5} where mountain shadows are confused with burnt land.

\begin{figure*}[!ht]
\centering
\includegraphics[width=18cm]{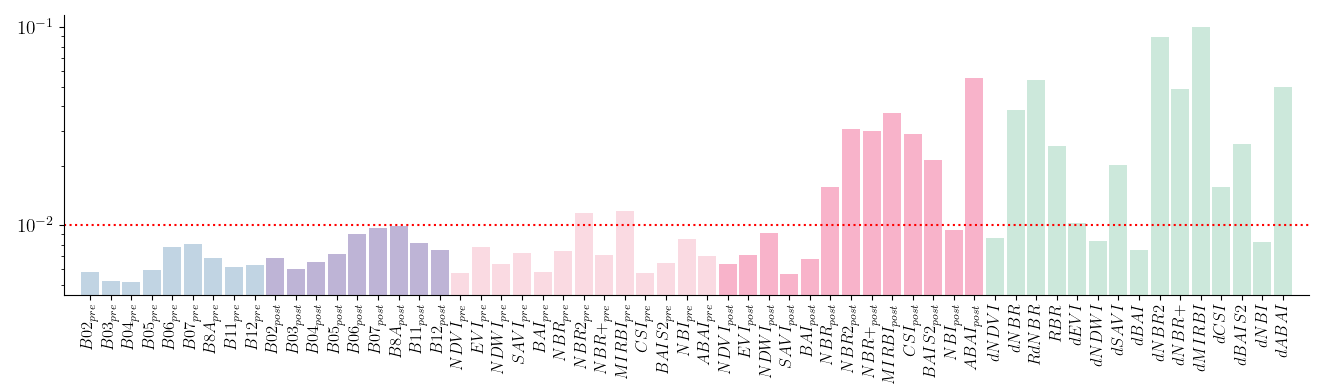}
\caption{Features importance as found by the RF classifier.}
\label{feature_import}
\end{figure*}

\subsection{Examination of ML methods}
\label{sec:ablation_ml}

\begin{figure}[!ht]
\centering
\includegraphics[width=\linewidth]{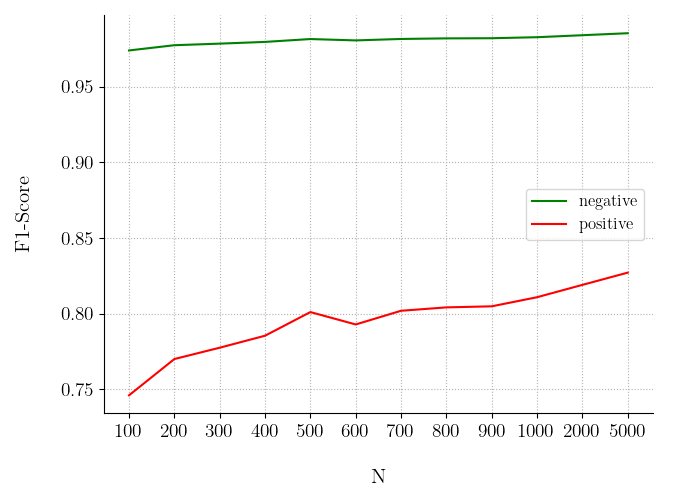}
\caption{F1-Score for the positive (burnt) and negative (unburnt) class in the validation set with respect to the hyperparameter $N$.}
\label{ml_datapoints}
\end{figure}

In order to assess the optimal value of the $N$ number of pixels to sample per patch, we performed grid search on the values $N$: \{100, 200, 300, 400, 500, 600, 700, 800, 900, 1000, 2000, 5000\}. We chose $N = 2000$ as it achieves the best F1 score for both the positive and negative class in the validation set (Fig. \ref{ml_datapoints}) and offers a good balance between accuracy and speed. Our final sub-dataset for the ML models consists of $763,173$ pixels in the training set and $276,726$ pixels in the validation set. The final evaluation is performed over the whole test set.

The next step is to reduce the great number of candidate input features and select a smaller subset so as to facilitate and speed up convergence. The candidate features include: all the band values for the pre- and post-fire image, the spectral indices presented in Tab. \ref{tab:indices} for pre- and post-fire images, as well as the differenced forms of the indices. First, we train an RF classifier on the whole set of input features and plot the feature significance (Fig. \ref{feature_import}). Based on this ordering, we conclude that the differenced indices and some specific bands of the post-fire image are found to contribute the most to the final prediction. In particular, bands falling in the Red-edge (B06), NIR (B07, B8A) and SWIR (B11, B12) regions display higher importance which is expected due to the clear spectral signature of burnt land in these bandwidths. As a next step, we proceed to create three different sub-datasets:

\begin{enumerate}
    \item{\textbf{Dataset All}: All features (bands, indices, differenced indices).}
    \item{\textbf{Dataset MI}: The features with importance $> 0.01$ as found by the RF classifier (see Fig. \ref{feature_import}).}
    \item{\textbf{Dataset dSI}: Only the differenced indices.}
\end{enumerate}

Finally, we deduce the optimal set of model hyperparameters by performing grid search over a number of candidate values. We repeat this process 5 times for each algorithm and end up with 4 different configurations achieving best results. Subsequently, we train and evaluate these selected models 5 times each in the different sub-datasets defined previously.

The final results are reported in Tab. \ref{tab:results_abl_ml}. We can see that the SVM trained on the differenced spectral indices is the best performing model, followed by RF trained on all input features. All ML models seem to suffer from high commission rates which is demonstrated by the high recall and low precision values in the burnt class. This can be mainly attributed to the misclassification errors produced by the spectral indices in the input which are then cascaded to the final predictions of the models.

An interesting takeaway is the fact that the set of input features that achieves the best results differs for each algorithm, i.e. MLP and RF perform best with all input features, whereas XGBoost and SVM benefit from the reduced subsets. This indicates that ML models require a careful and methodical feature engineering pipeline in order to improve performance and speed.

\begin{figure*}[!ht]
\centering
\includegraphics[width=7in]{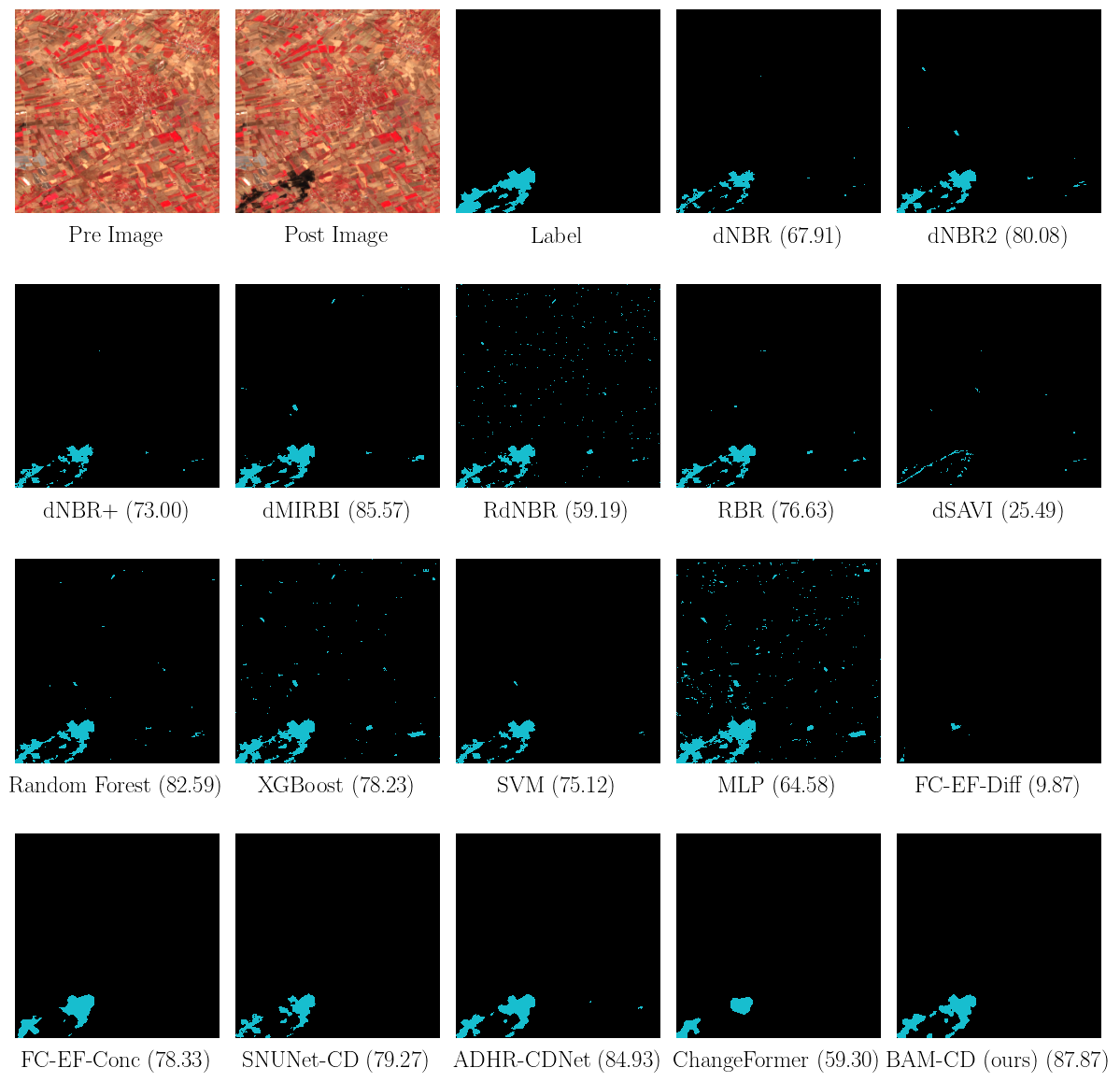}
\caption{Sample predictions on the test set. Numbers in parentheses indicate the F1-score. Satellite imagery is plotted as NIR-Red-Green composites.}
\label{test_vis_2}
\end{figure*}

\begin{figure*}[!ht]
\centering
\includegraphics[width=7in]{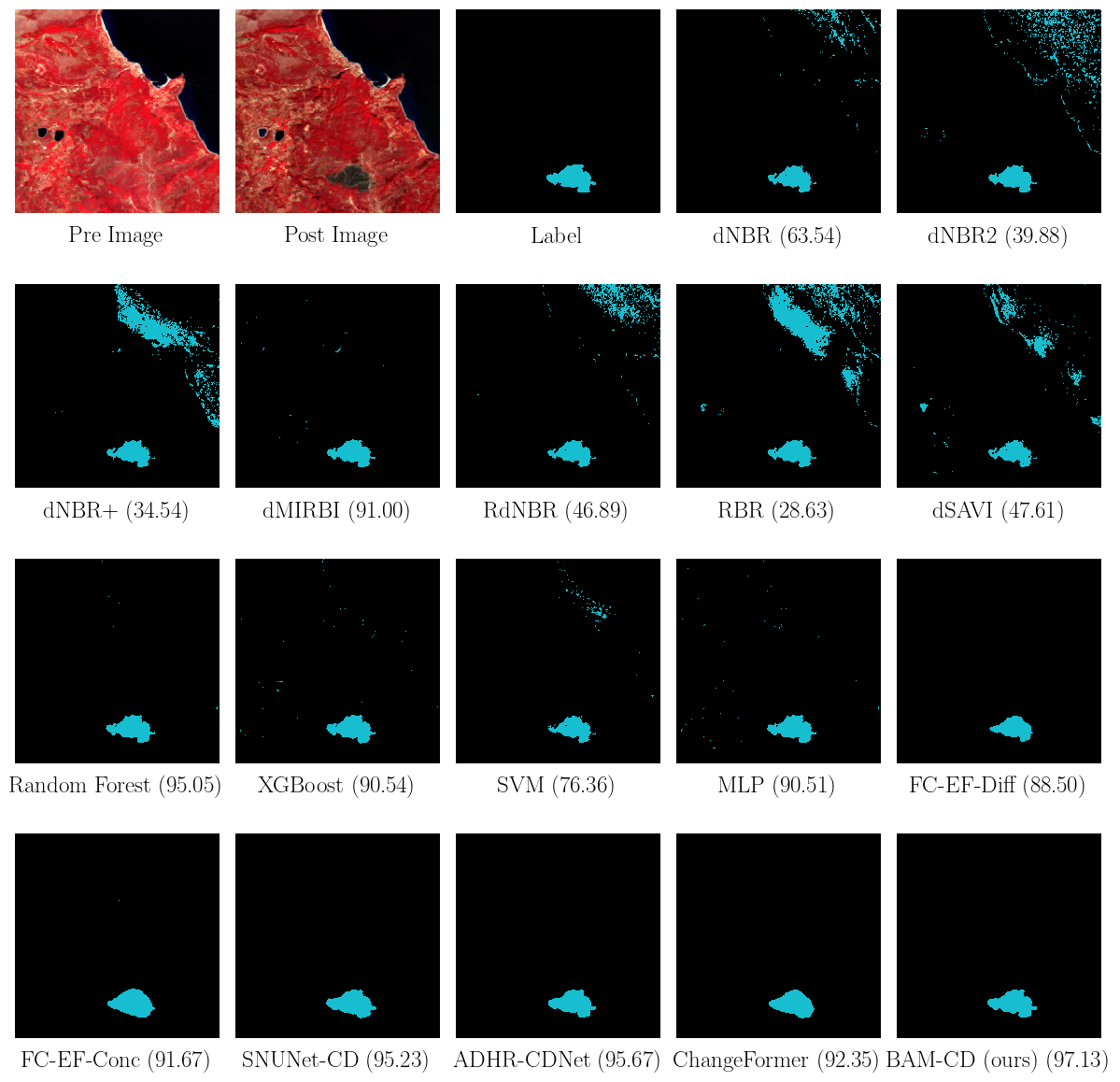}
\caption{Sample predictions on the test set. Numbers in parentheses indicate the F1-score. Satellite imagery is plotted as NIR-Red-Green composites.}
\label{test_vis_3}
\end{figure*}

\begin{figure*}[!ht]
\centering
\includegraphics[width=7in]{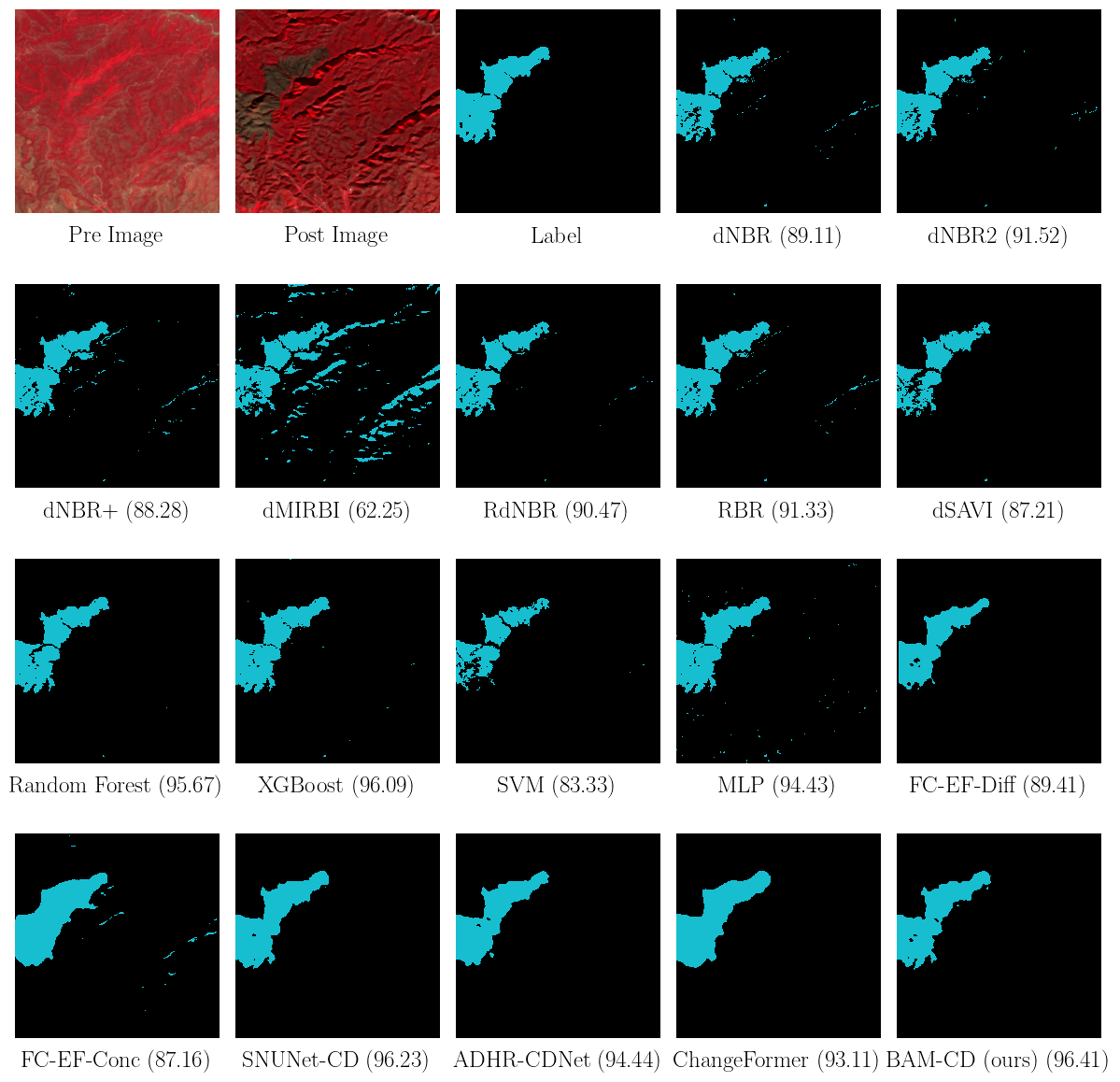}
\caption{Sample predictions on the test set. Numbers in parentheses indicate the F1-score. Satellite imagery is plotted as NIR-Red-Green composites.}
\label{test_vis_5}
\end{figure*}

In the visualized examples (Fig. \ref{test_vis_1}-\ref{test_vis_5}) we can see that MLP consistently produces a great number of false positive predictions, whereas SVM always underestimates the total burnt area, which is also reflected in the high precision and low recall scores in Tab. \ref{tab:results_abl_ml}. Fig. \ref{test_vis_2} is a solid demonstration of spectral confusion with cultivated land, where all four models make false predictions for burnt area on crops. Similarly, Fig. \ref{test_vis_3} displays confusion with sea, which however is more evident in the output of most spectral indices while the ML models manage to correctly classify sea as unburnt. In Fig. \ref{test_vis_5}, although most spectral indices tend to misclassify topography shadows as burnt land, the ML models correctly suppress this error. Finally, in Fig. \ref{test_vis_4} an older fire event which is visible in both images is marked in red. We can see that MLP cannot isolate the fire event which occurs between the images but is further confused by the older burnt land producing a great number of false alerts.

\subsection{BAM-CD ablation study}
\label{sec:ablation_dl}

In this set of experiments, we explore a number of variations and extensions of BAM-CD in order to assess their efficacy and usefulness in this particular task. First, we vary the model capacity by including backbones of different depth in the encoders (BAM-CD-r50 with a Resnet-50 architecture and BAM-CD-r101 with a Resnet-101). Subsequently, we add attention mechanisms \cite{roy2018recalibrating} in the decoder (Fig. \ref{modules_attn}) in order to help the model focus on important features and isolate the changes we are interested in (BAM-CD-r50-att and BAM-CD-r101-att respectively). All model variations are trained both in a Siamese (P) and a pseudo-Siamese (PS) scheme using the Binary Cross-Entropy loss (Eq. \ref{eq:bce}), and with the same optimizer and learning rate.

{\renewcommand{\arraystretch}{1.6}
\begin{table*}[ht!]
\caption{Results for the ML models ablation study. For each experiment we report the Precision (\%), Recall (\%), F1 Score (\%) and IoU (\%) for both classes as well as the mean F1 Score and the mean IoU. Numbers indicate the average value among all runs and values in parentheses indicate the standard deviation among all runs. Best values per model are in \textbf{bold}, second best are \underline{underlined}.}
\label{tab:results_abl_ml}
\centering
\hspace*{-1.2cm}
\tabcolsep=0.2cm
\setlength{\arrayrulewidth}{0.2mm}
\begin{tabular}{ |m{1.0cm}||m{1.2cm}||m{1.3cm} m{1.3cm} m{1.3cm} m{1.3cm}|m{1.3cm} m{1.3cm} m{1.3cm} m{1.3cm}|m{1.3cm} m{1.3cm}|| }
 \hline
  \multirow{2}{*}{\textbf{Model}} & \multirow{2}{*}{\textbf{Dataset}}
  & \multicolumn{4}{c|}{\textbf{Unburnt}} & \multicolumn{4}{c|}{\textbf{Burnt}} & \multicolumn{2}{c||}{\textbf{Total}} \\
  \multicolumn{1}{|c||}{} & \multicolumn{1}{c||}{} & \multicolumn{1}{c}{\textbf{Precision}} & \multicolumn{1}{c}{\textbf{Recall}} & \multicolumn{1}{c}{\textbf{F1}} & \multicolumn{1}{c|}{\textbf{IoU}} & \multicolumn{1}{c}{\textbf{Precision}} & \multicolumn{1}{c}{\textbf{Recall}} & \multicolumn{1}{c}{\textbf{F1}} & \multicolumn{1}{c|}{\textbf{IoU}}& \multicolumn{1}{c}{\textbf{F1}} & \multicolumn{1}{c||}{\textbf{IoU}} \\
  \hline
  \hline
   & All & 99.27 {\tiny (0.00)} & \textbf{97.52 {\tiny (0.01)}} & \textbf{98.39 {\tiny (0.00)}} & \textbf{96.83 {\tiny (0.01)}} & \textbf{57.12 {\tiny (0.07)}} & 82.13 {\tiny (0.02)} & \textbf{67.38 {\tiny (0.05)}} & \textbf{50.08 {\tiny (0.05)}} & \textbf{82.88 {\tiny (0.02)}} & \textbf{73.82 {\tiny (0.03)}} \\
   RF & MI & \underline{99.38 {\tiny (0.01)}} & \underline{96.87 {\tiny (0.01)}} & \underline{98.11 {\tiny (0.01)}} & \underline{96.29 {\tiny (0.01)}} & \underline{52.17 {\tiny (0.09)}} & \underline{85.03 {\tiny (0.34)}} & \underline{64.66 {\tiny (0.16)}} & \underline{47.78 {\tiny (0.17)}} & \underline{81.39 {\tiny (0.08)}} & \underline{72.03 {\tiny (0.09)}} \\
   & dSI & \textbf{99.45 {\tiny (0.00)}} & 95.76 {\tiny (0.01)} & 97.57 {\tiny (0.01)} & 95.25 {\tiny (0.01)} & 45.13 {\tiny (0.06)} & \textbf{86.85 {\tiny (0.01)}} & 59.39 {\tiny (0.05)} & 42.24 {\tiny (0.05)} & 78.48 {\tiny (0.03)} & 68.75 {\tiny (0.03)} \\
   \hline
    & All & 99.29 {\tiny (0.00)} & \textbf{96.34 {\tiny (0.00)}} & \underline{97.79 {\tiny (0.00)}} & \underline{95.68 {\tiny (0.00)}} & \underline{47.59 {\tiny (0.00)}} & 82.74 {\tiny (0.00)} & \underline{60.42 {\tiny (0.00)}} & \underline{43.29 {\tiny (0.00)}} & \underline{79.11 {\tiny (0.00)}} & \underline{69.48 {\tiny (0.00)}} \\
    XGBoost & MI & \underline{99.42 {\tiny (0.00)}} & \underline{96.24 {\tiny (0.00)}} & \textbf{97.80 {\tiny (0.00)}} & \textbf{95.70 {\tiny (0.00)}} & \textbf{47.88 {\tiny (0.00)}} & \underline{85.93 {\tiny (0.00)}} & \textbf{61.49 {\tiny (0.00)}} & \textbf{44.40 {\tiny (0.00)}} & \textbf{79.65 {\tiny (0.00)}} & \textbf{70.05 {\tiny (0.00)}} \\
    & dSI & \textbf{99.44 {\tiny (0.0)}} & 94.88 {\tiny (0.00)} & 97.11 {\tiny (0.00)} & 94.38 {\tiny (0.00)} & 40.54 {\tiny (0.00)} & \textbf{86.80 {\tiny (0.00)}} & 55.27 {\tiny (0.00)} & 38.12 {\tiny (0.00)} & 76.19 {\tiny (0.00)} & 66.28 {\tiny (0.00)} \\
   \hline
   & All & \textbf{99.10 {\tiny (0.00)}} & 96.23 {\tiny (0.00)} & 97.65 {\tiny (0.00)} & 95.4 {\tiny (0.00)} & 45.5 {\tiny (0.00)} & \textbf{78.32 {\tiny (0.00)}} & 57.56 {\tiny (0.00)} & 40.41 {\tiny (0.00)} & 77.61 {\tiny (0.00)} & 67.91 {\tiny (0.00)} \\
   SVM & MI & \underline{98.79 {\tiny (0.00)}} & \underline{98.12 {\tiny (0.00)}} & \underline{98.45 {\tiny (0.00)}} & \underline{96.95 {\tiny (0.00)}} & \underline{59.91 {\tiny (0.00)}} & \underline{70.07 {\tiny (0.00)}} & \underline{64.59 {\tiny (0.00)}} & \underline{47.7 {\tiny (0.00)}} & \underline{81.52 {\tiny (0.00)}} & \underline{72.33 {\tiny (0.00)}} \\
   & dSI & 98.73 {\tiny (0.00)} & \textbf{98.79 {\tiny (0.00)}} & \textbf{98.76 {\tiny (0.00)}} & \textbf{97.55 {\tiny (0.00)}} & \textbf{69.45 {\tiny (0.00}} & 68.42 {\tiny (0.00)} & \textbf{68.93 {\tiny (0.00)}} & \textbf{52.59 {\tiny (0.00)}} & \textbf{83.85 {\tiny (0.00)}} & \textbf{75.07 {\tiny (0.00)}} \\
   \hline
    & All & \underline{99.39 {\tiny (0.00)}} & \textbf{95.31 {\tiny (0.00)}} & \textbf{97.31 {\tiny (0.00)}} & \textbf{94.75 {\tiny (0.00)}} & \textbf{42.26 {\tiny (0.01)}} & 85.36 {\tiny (0.01)} & \textbf{56.52 {\tiny (0.01)}} & \textbf{39.4 {\tiny (0.01)}} & \textbf{76.91 {\tiny (0.00)}} & \textbf{67.08 {\tiny (0.00)}} \\
  MLP & MI & \underline{99.39 {\tiny (0.00)}} & \underline{93.85 {\tiny (0.01)}} & \textbf{96.54 {\tiny (0.00)}} & \underline{93.31 {\tiny (0.01)}} & \underline{36.04 {\tiny (0.02)}} & \underline{85.75 {\tiny (0.00)}} & \underline{50.71 {\tiny (0.02)}} & \underline{33.99 {\tiny (0.02)}} & \underline{73.62 {\tiny (0.01)}} & \underline{63.65 {\tiny (0.01)}} \\
   & dSI & \textbf{99.64 {\tiny (0.00)}} & 91.61 {\tiny (0.01)} & \underline{95.45 {\tiny (0.00)}} & 91.3 {\tiny (0.01)} & 30.63 {\tiny (0.02)} & \textbf{91.78 {\tiny (0.01)}} & 45.9 {\tiny (0.02)} & 29.8 {\tiny (0.01)} & 70.68 {\tiny (0.01)} & 60.55 {\tiny (0.01)} \\
 \hline
\end{tabular}
\vspace*{5mm}
\end{table*}
}

{\renewcommand{\arraystretch}{1.6}
\begin{table*}[ht!]
\caption{Results for the DL models ablation study. For each experiment we report the Precision (\%), Recall (\%), F1 Score (\%) and IoU (\%) for both classes as well as the mean F1 Score and the mean IoU. (S) Stands for a siamese and (PS) for a pseudo-siamese setting. Numbers indicate the average value among all runs and values in parentheses indicate the standard deviation among all runs. Best values are in \textbf{bold}, second best are \underline{underlined}.}\label{tab:results_abl}
\centering
\hspace*{-1.0cm}
\tabcolsep=0.2cm
\setlength{\arrayrulewidth}{0.2mm}
\begin{tabular}{ |m{3.1cm}||m{1.2cm} m{1.2cm} m{1.2cm} m{1.2cm}|m{1.2cm} m{1.2cm} m{1.2cm} m{1.2cm}|m{1.2cm} m{1.2cm}|| }
 \hline
  \multirow{2}{*}{\textbf{Model}}
  & \multicolumn{4}{c|}{\textbf{Unburnt}} & \multicolumn{4}{c|}{\textbf{Burnt}} & \multicolumn{2}{c||}{\textbf{Total}} \\
  \multicolumn{1}{|c||}{} & \multicolumn{1}{c}{\textbf{Precision}} & \multicolumn{1}{c}{\textbf{Recall}} & \multicolumn{1}{c}{\textbf{F1}} & \multicolumn{1}{c|}{\textbf{IoU}} & \multicolumn{1}{c}{\textbf{Precision}} & \multicolumn{1}{c}{\textbf{Recall}} & \multicolumn{1}{c}{\textbf{F1}} & \multicolumn{1}{c|}{\textbf{IoU}}& \multicolumn{1}{c}{\textbf{F1}} & \multicolumn{1}{c||}{\textbf{IoU}} \\
  \hline
  \hline
  BAM-CD-r50 (S) & 99.62 {\tiny (0.03)} & 99.25 {\tiny (0.1)} & 99.44 {\tiny (0.05)} & 98.88 {\tiny (0.1)} & 81.15 {\tiny (2.08)} & 89.63 {\tiny (0.83)} & 85.16 {\tiny (1.13)} & 74.17 {\tiny (1.7)} & 92.3 {\tiny (0.59)} & 86.53 {\tiny (0.9)} \\
  BAM-CD-r50 (PS) & 99.62 {\tiny (0.04)} & 99.46 {\tiny (0.09)} & 99.54 {\tiny (0.02)} & 99.09 {\tiny (0.05)} & 85.71 {\tiny (1.9)} & 89.56 {\tiny (1.22)} & 87.57 {\tiny (0.43)} & 77.89 {\tiny (0.67)} & 92.55 {\tiny (0.22)} & 88.49 {\tiny (0.36)} \\
  BAM-CD-r50-att (S) & \underline{99.68 {\tiny (0.04)}} & 99.44 {\tiny (0.05)} & \underline{99.56 {\tiny (0.02)}} & \underline{99.12 {\tiny (0.04)}} & 85.45 {\tiny (1.08)} & \underline{91.09 {\tiny (1.05)}} & \underline{88.17 {\tiny (0.43)}} & \underline{78.84 {\tiny (0.68)}} & \underline{93.86 {\tiny (0.22)}} & \underline{88.98 {\tiny (0.36)}} \\
  BAM-CD-r50-att (PS) & 99.59 {\tiny (0.06)} & \underline{99.53 {\tiny (0.06)}} & \underline{99.56 {\tiny (0.02)}} & \underline{99.12 {\tiny (0.03)}} & \underline{87.2 {\tiny (1.29)}} & 88.63 {\tiny (1.57)} & 87.89 {\tiny (0.43)} & 78.4 {\tiny (0.69)} & 93.72 {\tiny (0.22)} & 88.76 {\tiny (0.36)} \\
  BAM-CD-r101 (S) & 99.61 {\tiny (0.03)} & 99.4 {\tiny (0.02)} & 99.5 {\tiny (0.01)} & 99.01 {\tiny (0.02)} & 84.21 {\tiny (0.29)} & 89.13 {\tiny (0.8)} & 86.6 {\tiny (0.34)} & 76.37 {\tiny (0.53)} & 93.05 {\tiny (0.18)} & 87.69 {\tiny (0.28)} \\
  BAM-CD-r101 (PS) & 99.6 {\tiny (0.04)} & 99.46 {\tiny (0.06)} & 99.53 {\tiny (0.01)} & 99.07 {\tiny (0.02)} & 85.71 {\tiny (1.18)} & 88.82 {\tiny (1.2)} & 87.22 {\tiny (0.26)} & 77.34 {\tiny (0.41)} & 93.38 {\tiny (0.14)} & 88.2 {\tiny (0.22)} \\
  BAM-CD-r101-att (S) & \textbf{99.7 {\tiny (0.04)}} & 99.44 {\tiny (0.05)} & \textbf{99.57 {\tiny (0.01)}} & \textbf{99.14 {\tiny (0.01)}} & 85.63 {\tiny (0.97)} & \textbf{91.56 {\tiny (1.21)}} & \textbf{88.48 {\tiny (0.16)}} & \textbf{79.34 {\tiny (0.26)}} & \textbf{94.03 {\tiny (0.08)}} & \textbf{89.24 {\tiny (0.14)}} \\
  BAM-CD-r101-att (PS) & 99.52 {\tiny (0.04)} & \textbf{99.57 {\tiny (0.04)}} & 99.55 {\tiny (0.01)} & 99.1 {\tiny (0.01)} & \textbf{87.96 {\tiny (0.94)}} & 86.76 {\tiny (1.24)} & 87.35 {\tiny (0.19)} & 77.53 {\tiny (0.31)} & 93.45 {\tiny (0.1)} & 88.32 {\tiny (0.16)} \\
 \hline
\end{tabular}
\vspace*{5mm}
\end{table*}
}

Tab. \ref{tab:results_abl} shows the final results for the search of the optimal architecture. The best setup seems to be the Resnet-101 encoder trained with a Siamese scheme. Additionally, attention mechanisms seem to further improve the final results by helping the model focus on the most salient features extracted by the encoders.

Subsequently, we select the BAM-CD-r101-att (P) model, called BAM-CD from here onwards, and try out the most popular loss functions found in the literature for change detection settings: Binary Cross-entropy ($L_{BCE}$), Focal loss ($L_{Focal}$), Dice loss ($L_{Dice}$) and a sum of Binary Cross-entropy and Dice loss ($L_{BCE} + L_{Dice}$). Binary Cross-entropy (Eq. \ref{eq:bce}) is the traditional loss function for binary classification tasks, whereas Focal (Eq. \ref{eq:fl}) and Dice (Eq. \ref{eq:dl}) losses are most suitable for highly imbalanced class distributions. Apart from these common functions, a number of studies have adopted a sum of Binary Cross-entropy and Dice loss as the final objective function which seems to avert the exploding gradient phenomena inherent in Dice loss training.

\begin{equation}
\label{eq:bce}
L_{BCE}(y, \hat{y}) = -(y log(\hat{y}) + (1 -y)log(1 - \hat{y}))
\end{equation}

\begin{equation}
\label{eq:fl}
L_{Focal}(y, \hat{y}) = - \alpha_{t}(1 - p_{t})^{\gamma} log(p_{t})
\end{equation}

where $\alpha$ is a constant in $[0, 1]$ and $\gamma > 0$.

\begin{equation}
\label{eq:dl}
L_{Dice}(y, \hat{y}) = 1 - \frac{2y\hat{y} + 1}{y + \hat{y} +1}
\end{equation}

In all of the above equations, $y$ is the ground truth and $\hat{y}$ is the model prediction. As seen in Tab. \ref{tab:results_abl_loss}, the $L_{BCE}$ loss seems to yield the best metric scores, followed by the $L_{focal}$ loss. It is worth noting that $L_{Dice}$ and $L_{BCE} + L_{Dice}$ display a lot of instability as can be discerned from the higher standard deviation values, whereas higher accuracy and lower recall values indicate that they suffer less from false positive rates.
In any case, all other three loss functions do not seem as competitive alternatives to a simple Cross-entropy objective which achieves both higher F1 and IoU values. For that reason, we proceed by adopting $L_{BCE}$ for the subsequent experiments.

{\renewcommand{\arraystretch}{1.6}
\begin{table*}[ht!]
\caption{Results for the loss function ablation study using the BAM-CD model. For each experiment we report the Precision (\%), Recall (\%), F1 Score (\%) and IoU (\%) for both classes as well as the mean F1 Score and the mean IoU. Numbers indicate the average value among all runs and values in parentheses indicate the standard deviation among all runs. Best values are in \textbf{bold}, second best are \underline{underlined}.}\label{tab:results_abl_loss}
\centering
\hspace*{-1.0cm}
\tabcolsep=0.2cm
\setlength{\arrayrulewidth}{0.2mm}
\begin{tabular}{ |m{2.5cm}||m{1.3cm} m{1.3cm} m{1.3cm} m{1.3cm}|m{1.3cm} m{1.3cm} m{1.3cm} m{1.3cm}|m{1.3cm} m{1.3cm}|| }
 \hline
  \multirow{2}{*}{\textbf{Loss function}}
  & \multicolumn{4}{c|}{\textbf{Unburnt}} & \multicolumn{4}{c|}{\textbf{Burnt}} & \multicolumn{2}{c||}{\textbf{Total}} \\
  \multicolumn{1}{|c||}{} & \multicolumn{1}{c}{\textbf{Precision}} & \multicolumn{1}{c}{\textbf{Recall}} & \multicolumn{1}{c}{\textbf{F1}} & \multicolumn{1}{c|}{\textbf{IoU}} & \multicolumn{1}{c}{\textbf{Precision}} & \multicolumn{1}{c}{\textbf{Recall}} & \multicolumn{1}{c}{\textbf{F1}} & \multicolumn{1}{c|}{\textbf{IoU}}& \multicolumn{1}{c}{\textbf{F1}} & \multicolumn{1}{c||}{\textbf{IoU}} \\
  \hline
  \hline
  $L_{BCE}$ & \textbf{99.7 {\tiny (0.04)}} & 99.44 {\tiny (0.05)} & \textbf{99.57 {\tiny (0.01)}} & \textbf{99.14 {\tiny (0.01)}} & 85.63 {\tiny (0.97)} & \textbf{91.56 {\tiny (1.21)}} & \textbf{88.48 {\tiny (0.16)}} & \textbf{79.34 {\tiny (0.26)}} & \textbf{94.03 {\tiny (0.08)}} & \textbf{89.24 {\tiny (0.14)}} \\
  $L_{Focal}$ & \underline{99.64 {\tiny (0.03)}} & \underline{99.48 {\tiny (0.04)}} & \underline{99.56 {\tiny (0.01)}} & \underline{99.13 {\tiny (0.03)}} & 86.18 {\tiny (0.73)} & \underline{90.12 {\tiny (0.89)}} & \underline{88.1 {\tiny (0.38)}} & \underline{78.74 {\tiny (0.6)}} & \underline{93.83 {\tiny (0.2)}} & \underline{88.93 {\tiny (0.2)}} \\
  $L_{Dice}$ & 99.57 {\tiny (0.06)} & \textbf{99.51 {\tiny (0.09)}} & 99.54 {\tiny (0.02)} & 99.08 {\tiny (0.04)} & \underline{86.63 {\tiny (1.96)}} & 88.18 {\tiny (1.59)} & 87.36 {\tiny (0.43)} & 77.56 {\tiny (0.67)} & 93.45 {\tiny (0.22)} & 88.32 {\tiny (0.36)} \\
  $L_{BCE} + L_{Dice}$ & 99.59 {\tiny (0.04)} & \textbf{99.51 {\tiny (0.07)}} & 99.55 {\tiny (0.02)} & 99.1 {\tiny (0.04)} & \textbf{86.76 {\tiny (1.49)}} & 88.5 {\tiny (1.24)} & 87.6 {\tiny (0.47)} & 77.94 {\tiny (0.75)} & 93.58 {\tiny (0.25)} & 88.52 {\tiny (0.4)} \\
 \hline
\end{tabular}
\vspace*{5mm}
\end{table*}
}

\subsection{Comparison with other DL methods}
\label{sec:dl_comparison}

Tab. \ref{tab:results_all} summarizes the results of all the experiments. From the final metrics, we can safely assume that DL models consistently outperform more conventional ML approaches and spectral indices by a large margin, albeit at the expense of additional instability among runs.

Among the various DL techniques, SNUNet-CD achieves best results, proving that densely connected u-shaped architectures offer an additional advantage over self-attention mechanisms (ChangeFormer) and custom multi-scale approaches (ADHR-CDNet). As expected, simple U-Net architectures like FC-EF-Diff/Conc struggle to achieve performance competitive to more sophisticated models. Finally, our proposed architecture manages to outperform the rest by more than 1.01 \% on mean F1 score and 1.6 \% on mean IoU, which suggests that carefully designed deeper residual architectures offer a much better and targeted feature extraction than more complex approaches. Finally, the reported standard deviations achieved by BAM-CD are much lower than those of the other models, proving the stability and robustness of our approach.

\begin{figure*}[!ht]
\centering
\includegraphics[width=7in]{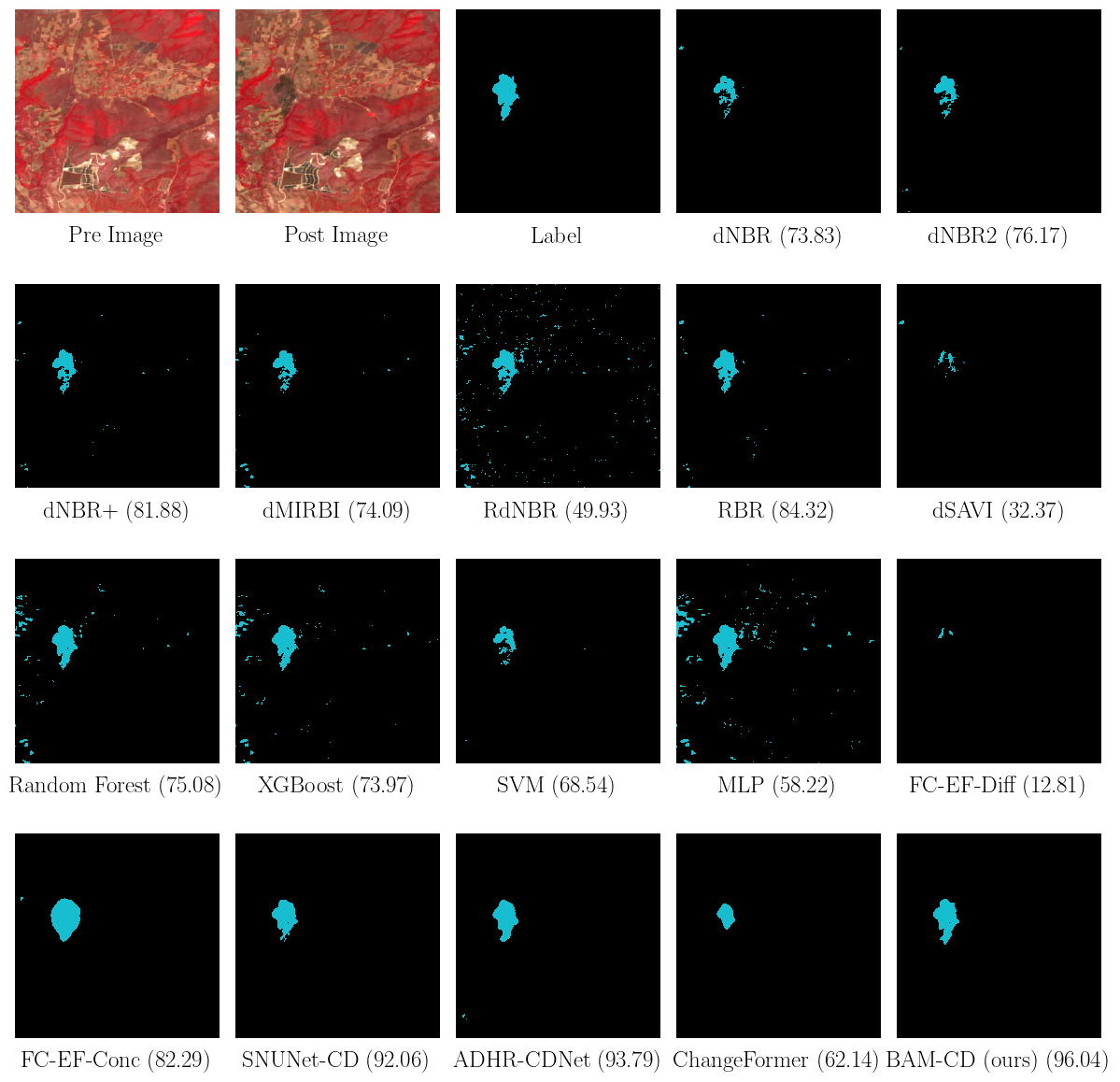}
\caption{Sample predictions on the test set. Numbers in parentheses indicate the F1-score. Satellite imagery is plotted as NIR-Red-Green composites.}
\label{test_vis_1}
\end{figure*}

\begin{figure*}[!ht]
\centering
\includegraphics[width=7in]{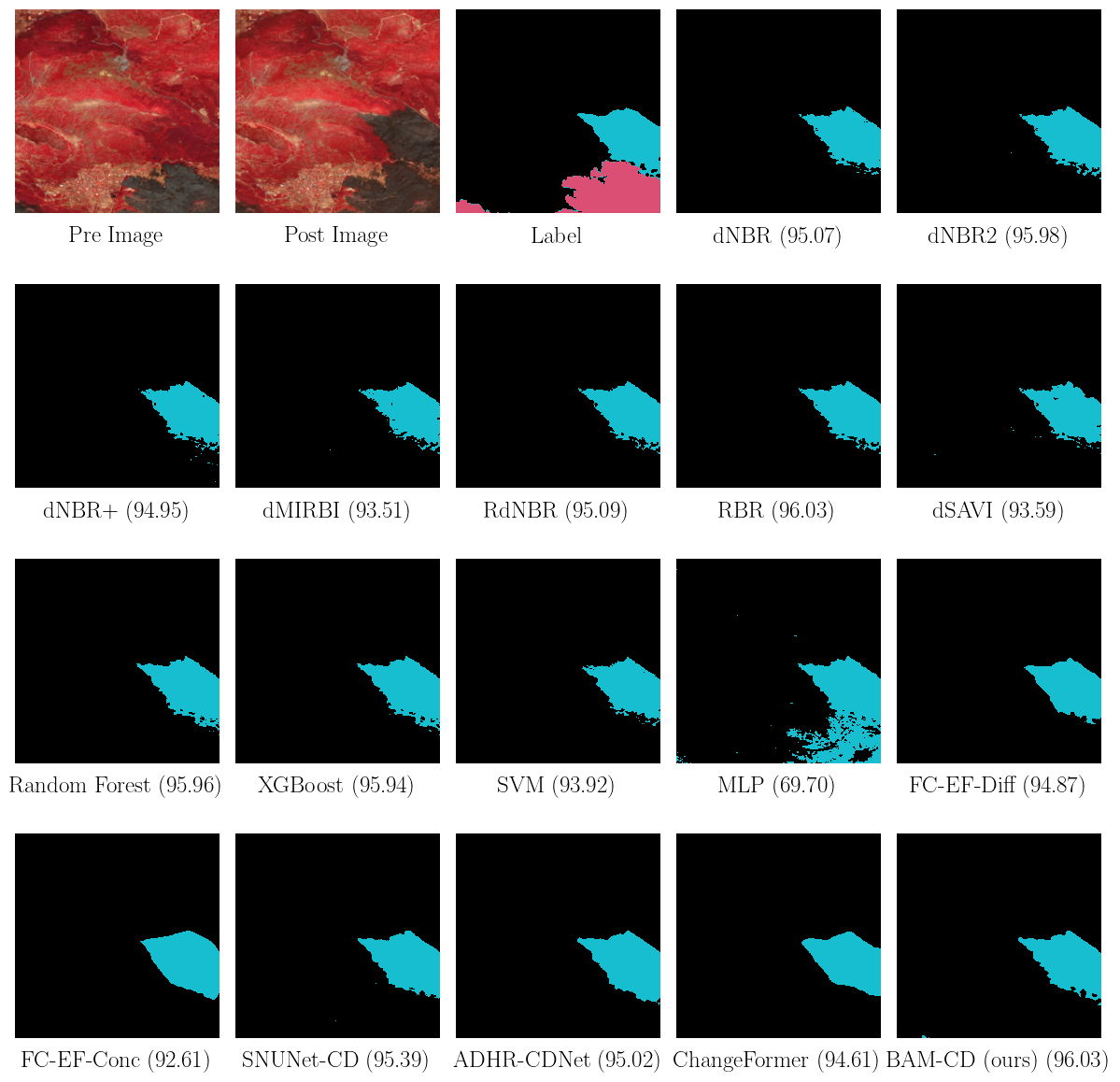}
\caption{Sample predictions on the test set. An older fire event in marked in red in the label. Numbers in parentheses indicate the F1-score. Satellite imagery is plotted as NIR-Red-Green composites.}
\label{test_vis_4}
\end{figure*}

Fig. \ref{test_vis_1} illustrates the high precision of our model. It manages to correctly capture the outline of the burn scar and does not get confused by nearby crop land or topography. The different spectral indices tend to underestimate the burnt area, while ML models display high false positive rate, esp. the MLP. FC-EF-Diff fails to discriminate the area of interest almost completely, and FC-EF-Conc predicts a rather coarse outline of the scar. ADHR-CDNet produces the second best prediction but the boundary of the scar is not as accurate. In Fig. \ref{test_vis_2} we can observe that DL models ignore crop land and vegetation and correctly focus on the fire event, as opposed to the spectral indices and ML models. ChangeFormer shows poorer performance than simpler algorithms and highly underestimates the burnt area. As can be seen in Fig. \ref{test_vis_3}, \ref{test_vis_5} and \ref{test_vis_4} DL models are also unaffected by the presence of sea/water, topography shadows or older burn scars, and manage to isolate the effects of the observed fire event.

\section{Discussion}
\label{sec:discussion}

Wildfires are among the most prevalent consequences of climate change and affect millions of people every year. Their destructive force and increased frequency pose a significant threat to ecosystems, wildlife and human communities. Therefore an accurate assessment of the caused damage through remote sensing data can provide an invaluable source of information for future protection and prevention measures as well as a guide for the successful restoration of the affected flora and fauna.

\begin{figure*}[!t]
\centering
\includegraphics[width=18cm]{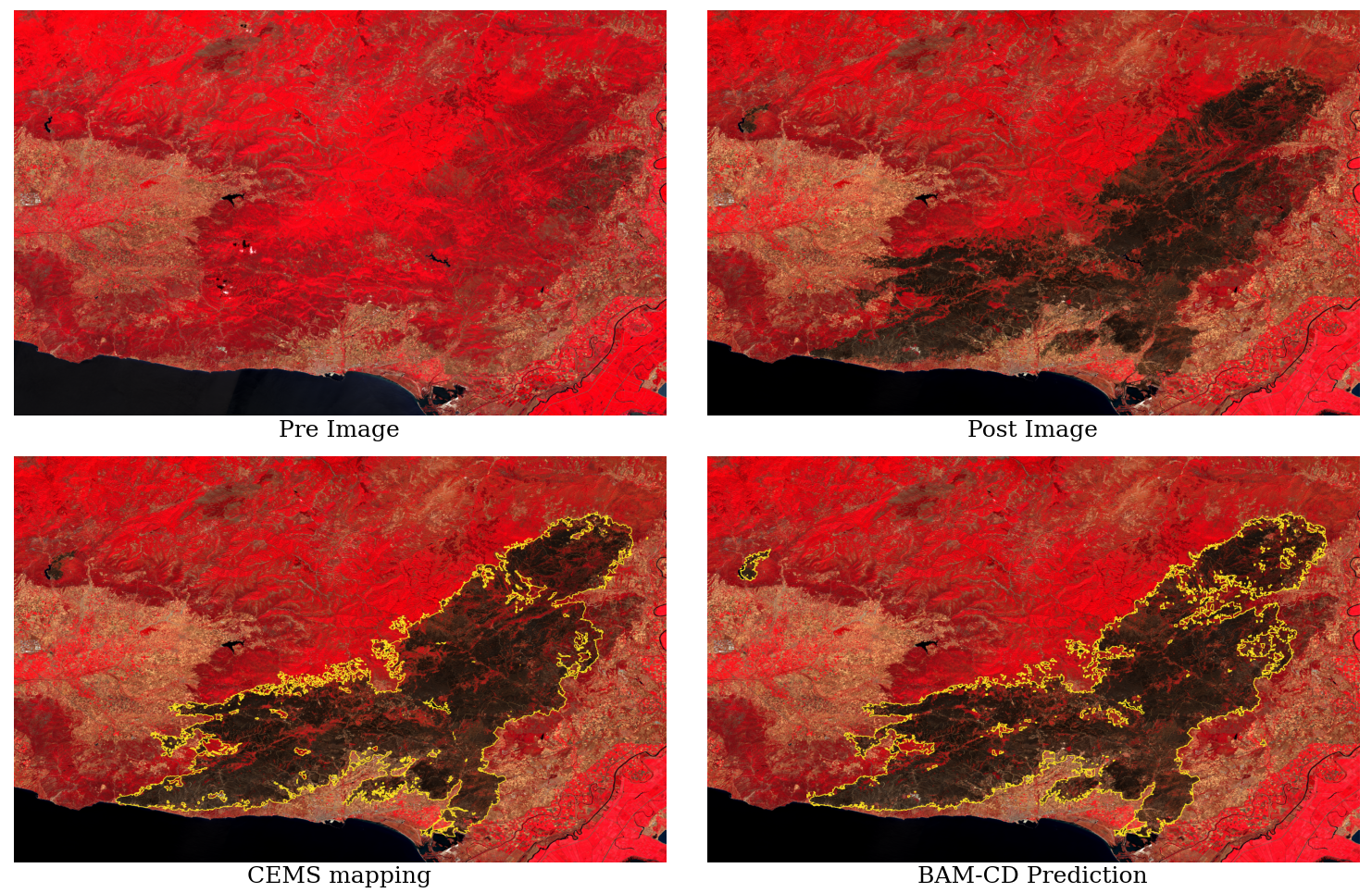}
\caption{Prediction of our model for the recent devastating wildfire at Evros, Greece (Aug. 2023). The official mapping of the Copernicus Emergency Management Service (CEMS) is also given as reference. Satellite imagery is plotted as NIR-Red-Green composites.}
\label{evros_pred}
\end{figure*}

To that end, we publish a novel, machine learning-ready dataset, namely FLOGA, which contains multi-source bitemporal satellite imagery provided by the Sentinel-2 and MODIS sensors. The ground truth labels cover several wildfire events in Greece for a span of 5 years. One of the most important traits of FLOGA is that the labeling process was executed by domain experts within the Hellenic Fire Service. These experts possess an intimate understanding of the intricacies and nuances associated with this specific task, as opposed to machine learning engineers or remote sensing scientists. Additionally, fire services and local authorities are the main end users of burn scar mapping systems, therefore the labels contained in FLOGA reflect their understanding of burnt area patterns, and needs and requirements in an operational setting, and set a solid target for any mapping method aimed to be adopted in real-world practical applications. We should note, however, that the labeling process was primarily based on satellite image photointerpretation without any field inspection in the areas of interest. This may have induced a certain bias in the data and small deviations from the ground truth may be possible.

With respect to the performance of the different models in the benchmark, the application of a number of spectral indices to the task of burnt area mapping has showed that the generalization of applicable thresholds to different land covers and imaging conditions is not straightforward. It was observed that spectral indices tend to get confused by surfaces exhibiting spectral signatures similar to burnt areas, such as water bodies, shadows, urban structures, etc. In addition, indices respond differently to the varying post-fire regeneration stages of different vegetation types and are highly sensitive to the time lag between the wildfire and the acquisition of the satellite image. Interestingly, indices utilizing the SWIR region of the e/m spectrum seem to perform best among all the candidate indices examined in this study.

Concerning the family of ML algorithms, the SVM trained on the differenced indices achieved the best results, surpassing RF by $\sim$0.97 in mean F1 score and $\sim$1.25 in mean IoU. However, visual inspection of the predicted binary maps show that all ML models suffer from severe ``salt n' pepper'' type of noise and high commission errors which were probably propagated by the spectral indices used in the input. Again, all ML approaches succumb to the same pitfalls as most indices when it comes to spectral confusion and sensitivity to imaging and atmospheric conditions.

On the other hand, DL models provide a robust alternative to the aforementioned traditional methods. In the present study, we examined two simple architectures (FC-EF-Diff/Conc) which were among the first ones to be proposed for change detection tasks, as well as three state-of-the-art models. As an additional contribution, we also proposed a novel DL architecture which manages to achieve best results both quantitatively and qualitatively. In general, all methods in this family produce more compact and accurate mappings of the burn scar and are not as prone to spectral confusion with similar surfaces. Our model, BAM-CD, surpasses spectral indices by $\sim$12.72 in mean F1 score and $\sim$17.1 in mean IoU, ML models by $\sim$10.18 in mean F1 score and $\sim$14.17 in mean IoU, and state-of-the-art DL models by $\sim$1.01 in mean F1 score and $\sim$1.6 in mean IoU.

In order to concretely establish the robustness of our model, we apply BAM-CD on a recent major wildfire event which ravaged the northeastern region of Evros, Greece, during August 2023. Considered the largest wildfire in Europe since the beginning of EFFIS records in 2000, it burned for 16 days and killed 20 people. According to the EU Copernicus service, the total burnt area rose to $93,880.8~ha$ ($938.81~km^2$).

We collected Sentinel-2 imagery on 29th July 2023 and 12th September 2023, ensuring that no image was contaminated with clouds, shadows and/or smoke. In Fig. \ref{evros_pred} both images along with the corresponding Copernicus Emergency Management Service (CEMS) official mapping and BAM-CD prediction are shown. A careful examination of the produced mapping displays the high precision of our model. The fire perimeter is adequately captured in high detail and unburnt spots inside the polygon are correctly excluded. The CEMS mapping is also given as reference. We can see that BAM-CD retrieves more unburnt islands than CEMS and a number of spot fires outside the fire perimeter are additionally mapped. It is also interesting to note that a different fire event located in the upper-left side of the images is also correctly mapped since it is absent in the first image acquisition and present in the second. Therefore, we have high expectations that the BAM-CD model can be effectively employed at an operational level, offering consistent support to government and forestry agencies for damage assessment and reforestation endeavors.

Despite the efforts of the authors to provide an in-depth analysis and comparison of many different approaches for the task of burnt area mapping, there is always room for improvement and further ideas. The proposed dataset covers the historic wildfire events in a single country, thus a crucial question is whether the explored techniques can be successfully generalized to different regions and climate zones. Especially in the case of Mediterranean ecosystems, due to similarities in biomes and climate conditions little effort is expected to be required for transferring the proposed methods. However, this remains an open research question and must be further examined. On the other hand, megafire events in completely different ecosystems, such as Canada and Australia, which are characterized by significantly different meteorological conditions, vegetation types and carbon stocks, may require a more elaborate pipeline for proper domain adaptation. For example, when labels in the target region are available, the training dataset could be enriched with examples of such events and further model finetuning or supervised domain adaptation techniques such as semantic alignment loss training \cite{motiian2017unified} could be applied. Furthermore, when labeled data are scarce or non-existent, semi-supervised (e.g. \cite{saito2019semi}, \cite{berthelot2021adamatch}) and unsupervised (e.g. \cite{ganin2015unsupervised}, \cite{hoffman2018cycada}) domain adaptation methods can be adopted respectively.

Future research could additionally explore alternative methods for the discovery of global thresholds for the spectral indices as well as innovative methods for the alleviation of spectral confusion. On top of that, a more exhaustive exploration on the feature engineering pipeline could greatly benefit the performance of the ML models, whereas alternative architectures, attention mechanisms (e.g. pyramid attention, transformers) and training schemes (e.g. generative, self-supervised) could also be examined for boosting the performance and accuracy of the proposed DL model. In addition, since BAM-CD is not particularly tailored for the task of burnt area mapping, but is rather a generic architecture, a possible application of the model to other change detection problems could be further examined. Finally, by leveraging both modalities provided in FLOGA (i.e. MODIS and Sentinel-2), one could possibly develop a downscaling approach for an accurate, near real-time high-resolution mapping with multi-resolution input data.

\section{Conclusion}
\label{sec:conclusion}

In this study we publish a novel analysis-ready dataset for the mapping of various wildfire events in Greece during the period 2017-2021. We include bitemporal multispectral satellite imagery of varying spatial resolution (MODIS and Sentinel-2), handcrafted labels provided by experts in the Hellenic Fire Service as well as auxiliary data such as land cover and water masks. We also perform an extensive benchmarking of the different methods proposed in literature for the task of burnt land mapping through a change detection approach and design a robust Deep Learning model that achieves superior performance. Even though DL has infiltrated many research domains and burnt area mapping seems to be an excellent fit, paradoxically there exist no studies to rigorously test different methods from Computer Vision to that extend and to our knowledge we are the first ones to attempt this comparison.

We hope that the contributions of this work will help propel further research in this area and provide the tools for a thorough exploration of viable solutions and approaches. We are confident that the conclusions of our study will assist scientists in their understanding of the strengths and shortcomings of the various techniques used in literature as well as operational settings.

\section*{Acknowledgments}

We would like to thank the Hellenic Fire Service for kindly providing the polygons of the burnt areas used in the proposed dataset. This work has received funding from the TREEADS project of the European Union’s Horizon 2020 research and innovation programme under grant agreement No 101036926.

\bibliographystyle{IEEEtran}
\bibliography{IEEEabrv,bibliography}

\clearpage

\clearpage

\clearpage

\clearpage

\clearpage

\end{document}